%% file: main_arxiv.tex
\documentclass[10pt,twocolumn,letterpaper]{article}

\usepackage[pagenumbers]{cvpr} % arXiv version with page numbers
\usepackage{booktabs}
\usepackage{mathtools}
\usepackage[accsupp]{axessibility}

\usepackage{listings}
\usepackage{xcolor}
\usepackage{multirow}
\usepackage{algorithm}
\usepackage{algpseudocode}
\usepackage{float}


\definecolor{cvprblue}{rgb}{0.21,0.49,0.74}

\newif\ifshownotes
\ifdefined\MakeWithNotes
  \shownotestrue
\fi
\ifdefined\MakeWithoutNotes
  \shownotesfalse
\fi

\ifshownotes
  \newcommand{\colornote}[3]{{\color{#1}\bf{#2: #3}\normalfont}}
  \newcommand{\colornoteTwo}[3]{{\color{#1}\bf{#3}\normalfont}}
  \newcommand{\colornoteThree}[2]{{\color{#1}\bf{#2}\normalfont}}
  \newcommand{\colornoteFour}[2]{{\color{#1}#2}}
\else
  \newcommand{\colornote}[3]{}
  \newcommand{\colornoteTwo}[3]{}
  \newcommand{\colornoteThree}[2]{}
  \newcommand{\colornoteFour}[2]{#2}
\fi

\definecolor{darkgreen}{rgb}{0.0,0.65,0}

\newcommand{\camready}[1]{\colornoteFour{orange}{#1}}

\input{macros}

\newcommand{\sref}[1]{\ref{#1}}

\usepackage[pagebackref,breaklinks,colorlinks,allcolors=cvprblue]{hyperref}

\def\paperID{} % *** Enter the Paper ID here
\def\confName{CVPR}
\def\confYear{2026}

\title{Self-Consistency for LLM-Based Motion Trajectory Generation and Verification}

\author{
Jiaju Ma\\Stanford University
\and
R. Kenny Jones\\Stanford University
\and
Jiajun Wu\\Stanford University
\and
Maneesh Agrawala\\Stanford University
}

\begin{document}

%% Suppress main paper sections from TOC; re-enabled before supplementary
\addtocontents{toc}{\protect\setcounter{tocdepth}{-1}}

%% teaser
\twocolumn[{%
\renewcommand\twocolumn[1][]{#1}%
\maketitle
\vspace{-0.8cm}
\begin{center}
    \centering
    \captionsetup{type=figure}
    \includegraphics[width=\textwidth]{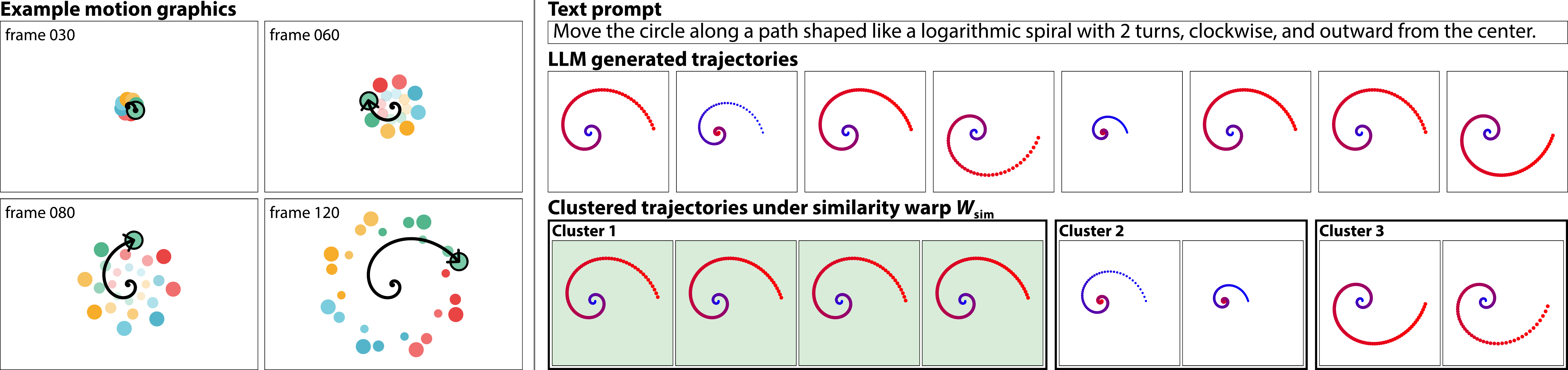}
    \vspace{-0.5cm}
    \captionof{figure}{Complex motion graphics animations are often composed of trajectories in the form of geometric shapes (left). While LLMs can generate motion graphics animations from a prompt describing the shape of an object's trajectory, the resulting animation does not always follow the prompt specification (right, motions move from blue to red). 
    We present a self-consistency method that enables more accurate LLM-based trajectory generation without supervision and show that it can be used for trajectory verification. We ask the LLM to generate multiple trajectory samples, cluster the samples using a hierarchy of geometric transformation groups, and choose the largest cluster as the most self-consistent set. We choose the centroid of the largest cluster as the most self-consistent generation, and verify a new trajectory by checking whether it can be added to the largest cluster (i.e. its distance to this centroid falls within a threshold $\tau$).
    }
    \label{fig:teaser}
\end{center}%
}]

\input{sec/abstract}    
\input{sec/introduction}
\input{sec/related_work}

\input{sec/preliminary}
\input{sec/method}
\input{sec/experiments}

\input{sec/discussion}

\input{sec/acknowledgments}

{
    \small
    \bibliographystyle{ieeenat_fullname}
    \bibliography{main}
}

%% ============================================================
%% SUPPLEMENTARY MATERIAL
%% ============================================================
\clearpage
\appendix

%% Reset counters for supplementary numbering
\renewcommand{\thesection}{\Alph{section}}
\setcounter{section}{0}
\renewcommand{\thefigure}{S\arabic{figure}}
\setcounter{figure}{0}
\renewcommand{\thetable}{S\arabic{table}}
\setcounter{table}{0}

%% Push floats to the top of float-only pages
\makeatletter
\setlength{\@fptop}{0pt}
\setlength{\@dblfptop}{0pt}
\makeatother

%% Re-enable TOC for supplementary sections
\addtocontents{toc}{\protect\setcounter{tocdepth}{3}}

\twocolumn[{%
\renewcommand\twocolumn[1][]{#1}%
\vspace{3em}
\maketitlesupplementary
\vspace{0.5em}
\begin{center}
    \centering
    \captionsetup{type=figure}
    \includegraphics[width=\textwidth]{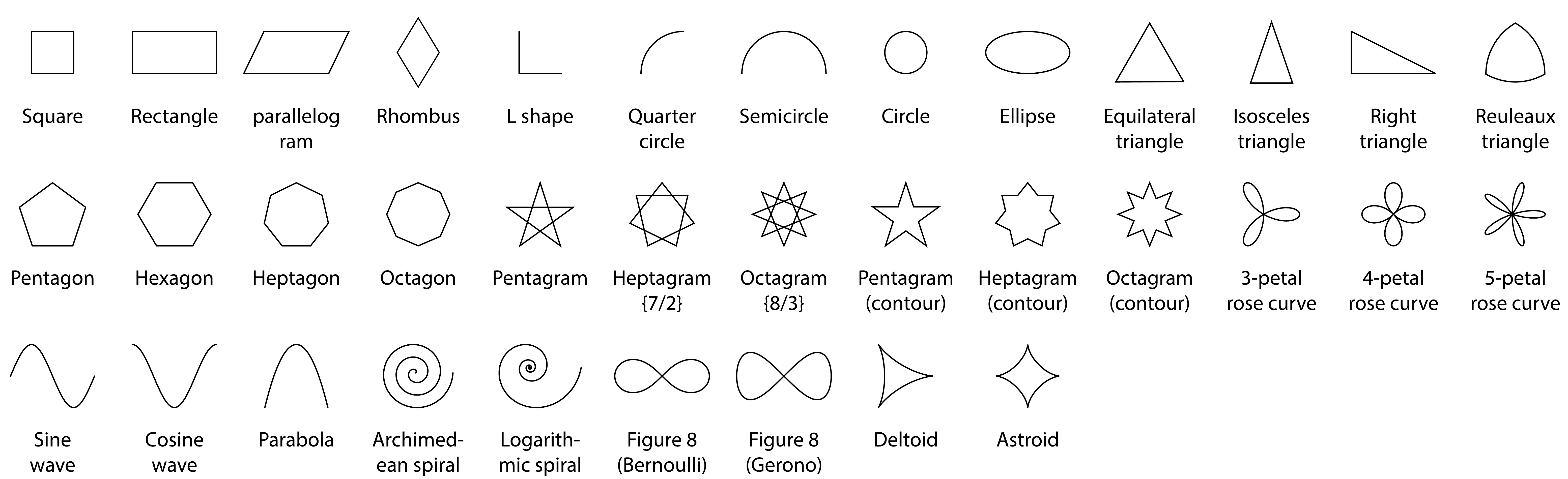}
    \caption{Our benchmark includes 224 prompts of motion trajectories derived from 35 different geometric base shapes. Variations are created by adding specifications to each shape to exercise different warps in our hierarchy of transformations.
    }
    \label{fig:benchmark_shapes}
\end{center}%
}]

\tableofcontents

\input{suppl_sec/benchmark_details}

\input{suppl_sec/experiment_details}

\input{suppl_sec/shape_family_limitation}
\input{suppl_sec/failure_modes}
\input{suppl_sec/method_details}
\input{suppl_sec/benchmark_figures}

\end{document}

%% file: macros.tex
\newcommand{\DCmajor}{\text{Majority-Consensus}}
\newcommand{\DChier}{\text{Hierarchical-Consistency}}
\newcommand{\DCmost}{\text{Most-Restrictive}}
\newcommand{\DCleast}{\text{Least-Restrictive}}

%% file: sec/abstract.tex
%%%%%%% DRAFT 3

\begin{abstract}
Self-consistency has proven to be an effective technique for improving LLM performance on natural language reasoning tasks in a lightweight, unsupervised manner. 
In this work, we study how to adapt self-consistency to visual domains.
Specifically, we consider the generation and verification of LLM-produced motion graphics trajectories. 
Given a prompt (e.g., ``Move the circle in a spiral path''), we first sample diverse motion trajectories from an LLM, and then identify groups of consistent trajectories via clustering. 
Our key insight is to model the family of shapes associated with a prompt as a prototype trajectory paired with a group of geometric transformations (e.g., rigid, similarity, and affine). 
Two trajectories can then be considered consistent if one can be transformed into the other under the warps allowable by the transformation group.
We propose an algorithm that automatically recovers a shape family, using hierarchical relationships between a set of candidate transformation groups. 
Our approach improves the accuracy of LLM-based trajectory generation by 4--6\%.
We further extend our method to support verification, observing 11\% precision gains over VLM baselines.
\camready{Our code and dataset are available at \url{https://majiaju.io/trajectory-self-consistency}.}
\end{abstract}

%% file: sec/introduction.tex
\section{Introduction}

Self-consistency has recently emerged as an effective method for improving the performance of LLMs on tasks with natural language outputs\,\cite{wang2022self}. 
At a high-level, the approach samples an LLM to produce diverse responses, and then groups responses to find the most consistent answer.
Consider a math problem like "A train travels 45 miles per hour. How far does it travel in 3 hours?" 
Self-consistency would use an LLM to sample diverse chain-of-thought reasoning traces, identify the final numerical value produced by each response, and return the most common number as the answer. 
This paradigm makes two base assumptions: (i) that the LLM is likely to produce the correct response more often than any other incorrect response; 
(ii) that it is possible to identify when answers are consistent with one another. 
Self-consistency is a useful paradigm, because it offers an unsupervised, training-free method for producing better generations from generative AI systems.

Beyond text domains, LLMs have also proven effective as generative systems for visual outputs, often via program synthesis. These models take a text prompt as input and generate a program (often in a domain specific language) that is then executed to render a visual image\,\cite{gupta2023visprog}, vector graphics\,\cite{polaczek2025neuralsvg,xing2024llm4svg}, 3D scenes\,\cite{hu2025scenecraft,zhang2024scenelanguage}, or animation\,\cite{gal2024breathing,liu2024logomotion,tseng2025keyframer,ma2025mover}. 
%This is commonly done by generating a program, or structured representation, which can then be converted into native visual data by a combination of execution and rendering engines. 
This leads to a question: how can we apply self-consistency approaches in the context of LLM-based visual generation?
If we want to identify the most common answer from a set of LLM responses, we need a way of checking whether different visual outputs are consistent with one another. 
 
Here we consider motion graphics animations as a motivating visual domain.
In such animations, it is common to describe the trajectories of SVG elements as forming different shapes such as parabolas, figure 8's, or spirals (Figure\,\ref{fig:teaser}).  
Given a prompt that describes a desired motion trajectory (e.g., ``Move the circle in a logarithmic spiral path''), we can ask an LLM to produce an SVG motion program that renders to a corresponding animation. 
By incorporating prompting strategies that encourage diversity\,\cite{zhang2025verbalized}, we can use an LLM to produce representative samples of motion trajectories corresponding to the prompt.

To apply self-consistency to the resulting motion trajectories, we need to know which of the trajectories are consistent with one another -- this is a clustering problem.
For language-based reasoning tasks, self-consistency approaches often assume that this clustering step is trivially done through direct comparisons (e.g., identity matching of numerical values in math domains). 
%\maneesh{I think this is also done using l2 norms on numerical values -- should check -- probably also using embedding distances.} \kenny{Maybe good to discuss -- I don't think anyone has done this kind of aggregation at answer-level, though it is commonly done at reasoning trace level to form weighted votes for majority decision (see the new rel work section I added). As such, we may want to claim this as a contribution that could be useful more generally in the discussion.  } \maneesh{I think a sentence or two in the discussion could be good. Otherwise let's just comment out this set of comments here.}
%
For visual domains like motion trajectories, it is unlikely that any two trajectories will identically match (e.g., at a pixel-level). 
This is partly due to underspecification in the natural language prompt. A prompt like ``move the circle in a logarithmic spiral path'' does not describe a single motion trajectory, but rather a family of shapes.
%So if its not meaningful to use identity matching to compare two trajectories, how do we know when they are consistent with one another?
%
%If we consider the previous motivating prompt, "move in a parabolic path", notice that this prompt doesn't designate a single motion trajectory, it maps to a distribution \maneesh{ would use the term "family" here instead of distribution} of shapes.
One way to model the family of trajectories induced by a natural language prompt is as 
%a shape family.
%We choose to model shape families as 
a prototype trajectory together with a group of geometric transformations (e.g., rigid, similarity, affine, etc.): warps produced from this group specify the allowable transformations that preserve the semantic identity of the desired shape. 
%Importantly, the appropriate invariances (e.g., preserves angles, orientations, parallel lines, etc.) for each transformation group depend on the desired shape family. 

%\maneesh{Probably need a concrete example here (or perhaps two to explain differences between two families). I might use a more realistic example than the parabola -- probably a spiral as that can also illustrate the notion of reflections. I think introducing the families in terms of rigid, similarity, etc. might be useful here as it will get reader thinking in the right way.}
% For instance, the shape family of parabolas would admit only X transforms, whereas figure-8 trajectories may allow X transforms.

These transformation groups provide us with the machinery to perform the clustering required by self-consistency.
Specifically, two trajectories are consistent if one can be transformed into the other under the warps allowable by the transformation group.
This consistency property can be checked with a distance-based clustering algorithm, assuming we have access to a distance metric that is invariant under the transformation group for the shape family.
Thus, to apply self-consistency over a set of LLM generated motion trajectory samples, if we know the correct transformation group, we can cluster the sampled trajectories under the corresponding distance metric, and choose any response from the largest cluster.

In our case, however, the {\em correct} family and transformation group associated with the shape described in the input prompt are not known a priori.  
Our approach is to first consider a hierarchy of geometric shape families based on Lie transformation groups that expose different degrees of freedom in how the prototype is allowed to vary.
Our hierarchy is based on shape families commonly used to describe trajectories\,\cite{sable2022language}.
%\kenny{Maybe something like:  we choose this set of warps so that they are useful for shape families that people commonly describe trajectories with~\cite{sable2022language}. }
We cluster the LLM-produced motion trajectory samples using each transformation group in the hierarchy and we
%progressively from the most-to-least restrictive distance metrics;
propose two decision criteria that analyze properties of these clusterings to identify the geometric shape family (with an associated prototype and transformation group) 
that is most consistent with the LLM samples.

Once we know the shape family, we can choose any LLM-generated sample from the largest cluster as a self-consistent generation. 
We also show how to extend this self-consistency approach to verify that any new query trajectory matches the prompt. 
More specifically, we check whether the query trajectory is a member of the shape family identified using the self-consistency approach, by testing whether it can be added to the largest cluster for the family (e.g., its distance to the prototype is below a threshold).

%Clustering trajectories under this distance metric allows us to not only select the most self-consistent LLM production, it also provides a route for verification.
%We develop a procedure that uses this clustering strategy to produce a shape family, i.e. a prototype trajectory and a warp function.
%We define a hierarchy of warping functions (with associated distance metrics) that expose different degrees of freedom in how the prototype is allowed to vary.
%We can then cluster the LLM produced motion trajectories progressively from the most-to-least restrictive distance metrics;
%we propose various decision criteria that analyze properties of this clustering to choose a prototype and warp function that is most consistent with the LLM generations.
%Once we have a shape family, verification can be performed with a membership check that determines whether or not a new query trajectory is a member of this shape family.

To evaluate our approach, we design a benchmark set of over 200 prompts describing shapes common to motion graphics animations.
For generation, we find that our proposed extension of self-consistency over motion trajectories is able to improve the accuracy of LLM generations on this benchmark by 4-6\% for the unsupervised setting when the transformation group needs to be heuristically estimated.
Additionally, we adapt this benchmark to evaluate verification abilities by collecting over 2000 pairs of (trajectory, prompt) with labels indicating whether these match or mismatch with one another.
We find that our proposed unsupervised verification strategy, which uses self-consistency to automatically recover shape families, outperforms the alternative of performing verification with a VLM with precision increases of 11.8\% and F1 increases of 5.6\%. 

%% file: sec/related_work.tex
\section{Related Work}
\label{sec:related}

\paragraph{Self-consistency.}

Self-consistency~\cite{wang2022self} was originally introduced as a strategy for improving LLM decoding.
In the vanilla set-up, this is done by sampling diverse chains of thought, and then selecting a final answer through a majority vote, where aggregation is performed through identity matching over discrete outputs (numerical values or strings).
Subsequent work has expanded this idea by generalizing the unweighted majority vote to take into account consistencies across reasoning traces~\cite{jiang2025representationconsistencyaccuratecoherent}.
Semantic Self-Consistency \cite{knappe2025semanticselfconsistencyenhancinglanguage} clusters embeddings of reasoning traces,  filtering out outlier responses, and then performs a majority vote over the remaining answers.
Latent Self-Consistency \cite{oh2025latentselfconsistencyreliablemajorityset} similarly clusters embeddings of reasoning traces, does aggregation within this reasoning space, and then returns the answer from this reasoning chain.
Collectively, these works demonstrate that self-consistency benefits from more comprehensive aggregation strategies, although their notions of clustering are still limited to bins defined by discrete identity matches.
In order to extend self-consistency to complex visual domains, like families of shapes commonly observed in the trajectories of elements within motion graphics, we propose an approach that determines sample-level consistency under a flexible distance metric.

% Vanilla self-consistency: sample then marginalize. Aggregation can be done with majority vote (unweighted agg), or taking models probability of correct answers into account (weighted agg)

% Some approaches like "Representation Consistency for Accurate and Coherent LLM Answer Aggregation" have looked at better ways of using the models internal activations to do this weighting over the answers, but ultimately still a weighted majority vote on identity matching.

% For text-based domains, a few other approaches have also looked into slightly more general notions of aggregation.

% One idea here is to try to identify similiar/disimiliar reasoning paths.

% "Semantic Self-Consistency: Enhancing Language Model Reasoning via Semantic Weighting”, they convert reasoning traces into embeddings, cluster to identify outliers, then do majority vote on answers from this filtered set.

% In the same general vein, "Latent Self-Consistency for Reliable Majority-Set Selection in Short- and
% Long-Answer Reasoning" embeds reasoning traces, uses clustering to find the most representative reasoning trace, and then chooses the answer from this response (aggregation is done only over reasoning traces, not final answers).

% \subsection{Visual program synthesis}
% - viperGPT, VisProg
% - scenecraft
% - self-consistency

\paragraph{Visual content verification.}

As generative AI systems have been used to create visual content across domains, recent work has developed various verification methods to ensure generated outputs match input prompts in terms of satisfying desired properties~\cite{hu2023tifa,sun2024dreamsync, si2025design2code,hu2025scenecraft,torre2024llmr}.
For image generation, many works employ a VQA-based verification paradigm.
TIFA converts image generation prompts into question-answer pairs that a VQA model can verify~\cite{hu2023tifa}. 
DSG ensures the validity of these questions by structuring them as dependency graphs~\cite{cho2024davidsonian}.
DreamSync applies verification in an iterative optimization loop, using VLM feedback to refine generations until they align with the prompt~\cite{sun2024dreamsync}.
Verification strategies have also been proposed for other visual modalities.
Design2Code proposes automatic metrics for evaluating LLM-generated webpages against design specifications~\cite{si2025design2code}.
SceneCraft uses LLM-based question answering to verify 3D Blender scene generation~\cite{hu2025scenecraft}, while LLMR applies analogous techniques to VR scene synthesis~\cite{torre2024llmr}.
VPGen demonstrates that synthesizing verification programs can provide more interpretable verification results than pure VLM evaluation~\cite{cho2023vpt2i}.

The work most closely related to ours is MoVer~\cite{ma2025mover}, which introduces a first-order logic DSL for verifying motion graphics animations.
While this approach enables verification of fine-grained spatio-temporal properties such as motion attributes and timing relationships, it requires animations to be explicitly specified in terms of low-level motion primitives and atomic predicates.
More importantly, it cannot effectively express the geometric shape families that people commonly use to describe motion trajectories.
In our work, we propose a set of self-consistency-based methods to enable the verification of motion trajectories.

%% file: sec/preliminary.tex
\begin{figure}[t]
    \centering
    \includegraphics[width=\linewidth]{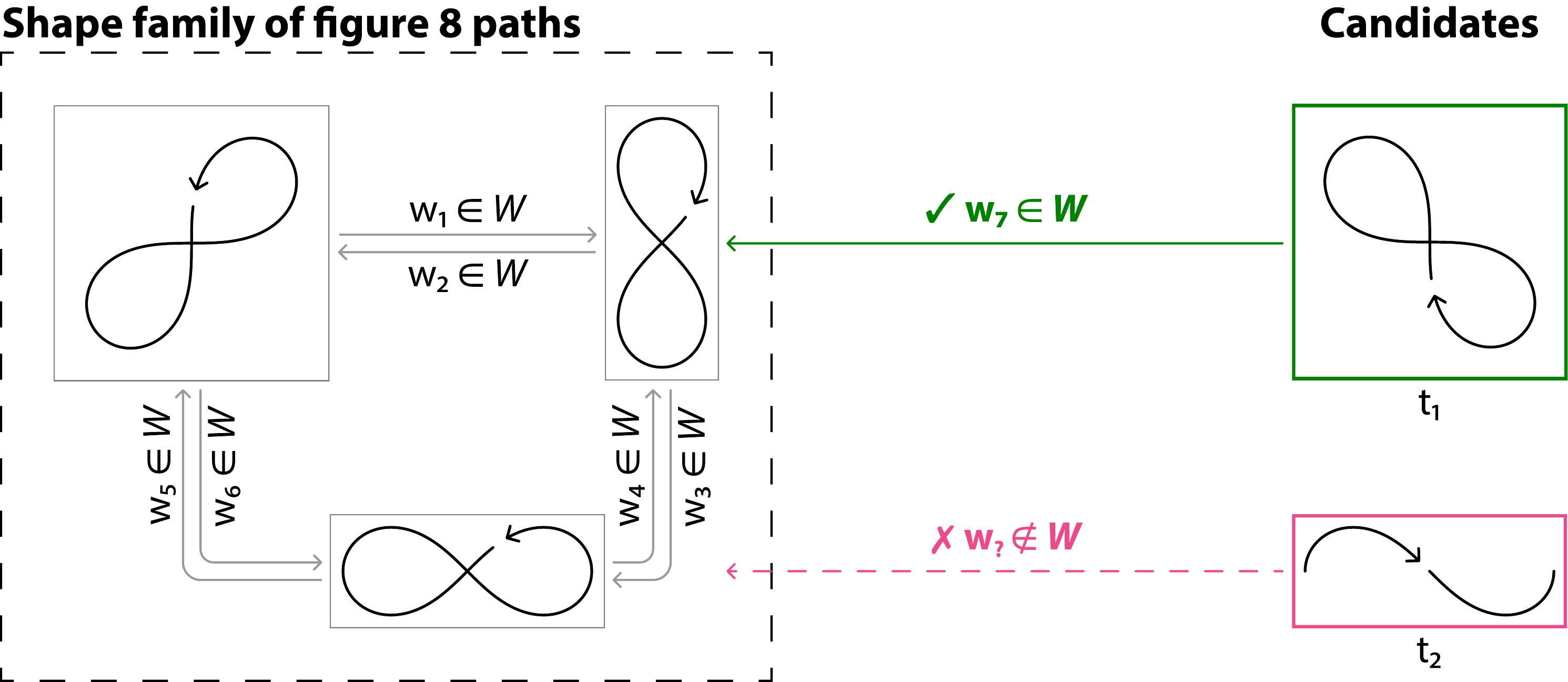}
    \caption{%
    We define a shape family $\mathcal{F}(o,W)$ with a prototype trajectory $o$ and a Lie transformation group $W$.
    The shape family is then the set of all trajectories that can be warped into one another $w(o)$ by all warps $w \in W$ (left).
    In this example, any of the figure 8's in the family can serve as the prototype since they can all be warped into one another.
    We can check whether a trajectory $t_i$ is a member of the family by checking if there is a warp $w \in W$ that can transform $t_i$ into $o$ (right).
    % \vspace{5mm}
    % \maneesh{Update figure to label $t$ for the two cases on the right.}
    % \jiaju{$t_1 has w_7, t_2$ has no w}
    }
    \label{fig:family}
\end{figure}
\section{Preliminaries}
\label{sec:preliminary}

% \maneesh{Should introduce the idea of shape families earlier -- before this section or as introductory material for this section. Should outline the assumptions of our current algorithm -- i.e. prompt describes a mathematically well defined family of shapes -- well defined in terms of a warp the shape is invariant to. Will need to handle the nonuniform scale cases carefully when we describe such families. Probably need to explain the idea of prototypes in that section as well.}

% We begin by defining terms relevant to our self-consistency-based generation and verification methods.
%\kenny{If time, maybe have motivating sentence: We provide the necessary foundational definitions for our approach}

\paragraph{Motion trajectory.} In motion graphics, a graphical element (e.g., a circle, rectangle, triangle, etc.) is animated by transformations as a function of time.
We define the element's \textit{trajectory} as its center point position over time.
As shown in Figure~\ref{fig:teaser}, motion trajectories 
commonly found in various motion graphics animations often fall into geometrically well-defined shape families.
%
%\maneesh{Need to say somewhere -- probably after we describe the preliminaries that --
%in this work we will assume that prompts specify motion trajectories as shapes that fall into mathematically well-defined families in the sense that they can be represented by a single prototype and warp $W$} 
%\kenny{This is in method, so probably covered?}

\paragraph{Shape family.} 
%for a prompt that describes a motion trajectory in the form of a geometric shape, 
Informed by the Erlangen program~\cite{kisil2012erlangen} and Lie transformation groups~\cite{kaji2016concise, eade2014lie}, we define a shape family $\mathcal{F}(o, W)$
as the set of trajectories that can be generated by transforming a prototype trajectory $o$ 
using any warp $w$ in a transformation group $W$ (Figure~\ref{fig:family}). That is,
%as the set of  trajectories that can be generated from a prototype trajectory $o$ under a certain type of geometric warp $W$:
\begin{equation}
\label{eq:shape_fam}
\mathcal{F}(o, W) = \{w(o) \mid w \in W\}
\end{equation}
where the warp $w$ applies transformations such as translation, rotation, scaling, and shear to the prototype $o$.
For example, a text prompt such as \textit{``Move in a figure 8-shaped path''} corresponds to a shape family where $o$ is a figure 8 shape (lemniscate of Bernoulli) at any scale, position, orientation, or reflection. 
Thus $W$ is the Lie 
group of similarity transformations with reflections or $\mathrm{Sim}(2)$\,\cite{kaji2016concise, eade2014lie}. A warp $w$ in this group can apply any combination of translation, rotation, uniform scale, and reflection to $o$ to generate the trajectories of the shape family (Figure~\ref{fig:family}). 
%\maneesh{Maybe switch figure 8 example to a spiral in Fig 3}

\begin{figure}[t]
    \centering
    \includegraphics[width=\linewidth]{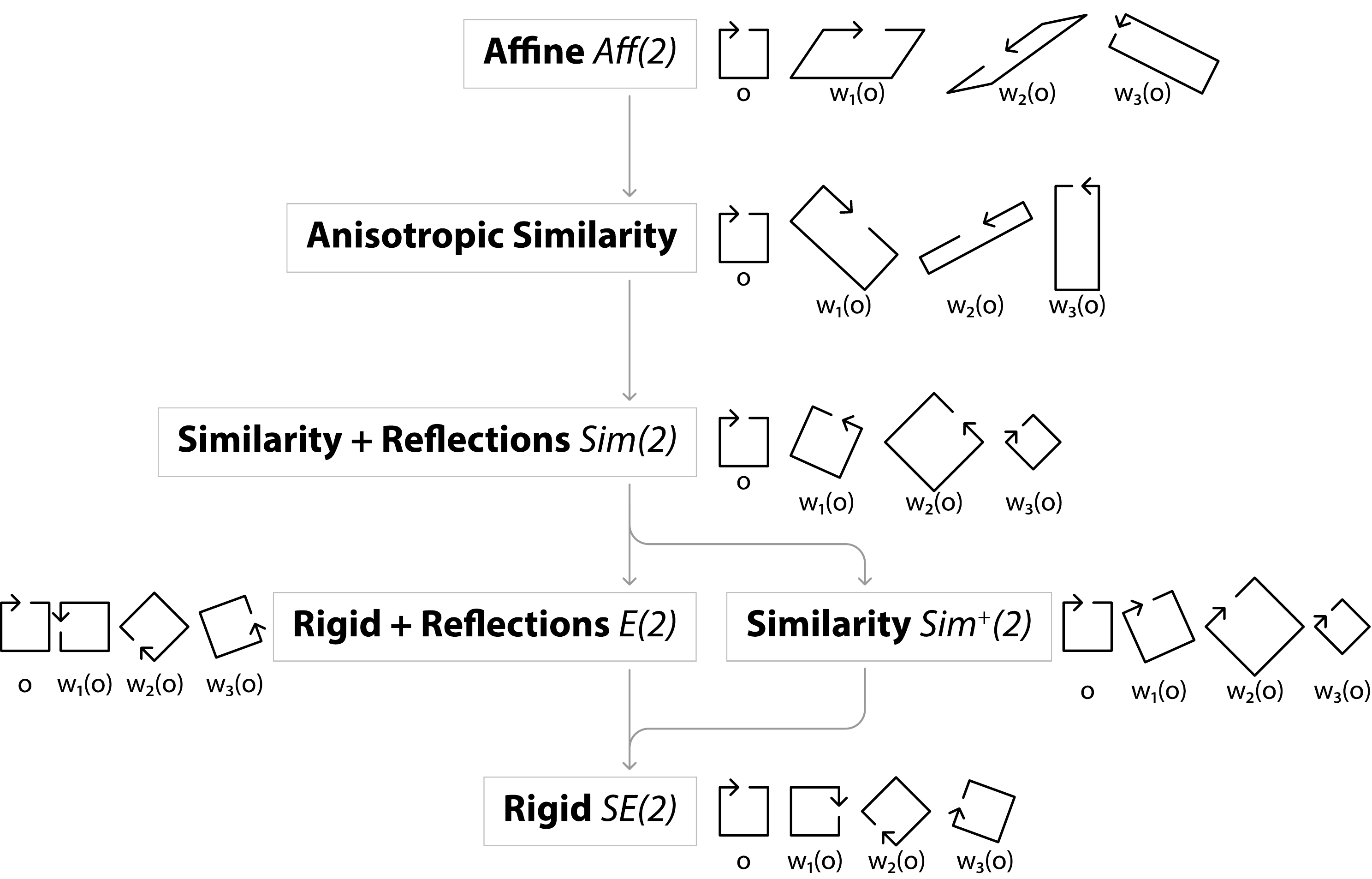}
    \caption{Hierarchy of Lie transformation groups and shape families they induce. Each node represents a  transformation group and depicts 
    a prototype square-shaped trajectory $o$ as well as other trajectories $w(o)$ within the corresponding shape family.
    % \maneesh{label with SE(2) labels. Label the prototype trajectory $o$ and maybe the others as $w_1(o), w_2(0), ...$}
    % \jiaju{tiny labels}
    }
    % \vspace{5mm}
    \label{fig:hierarchy}
\end{figure}

In this work we develop shape families using a set of Lie transformation groups that people commonly consider when describing a family of shapes\,\cite{sable2022language}.

\vspace{2mm}
\noindent
\textbf{\textit{Rigid}} $[W_{\text{rgd}} \coloneq \mathrm{SE}(2)]$:
The special Euclidean group that includes all combinations of translations and rotations. 
This group preserves all distances, angles, and orientation. Note that while this group is commonly denoted $\mathrm{SE}(2)$ it is sometimes also denoted $\mathrm{E}^+(2)$.

\vspace{2mm}
\noindent
\textbf{\textit{Rigid + reflections}} $[W_{\text{rgd-ref}} \coloneq \mathrm{E}(2)]$:
The full Euclidean group is the set of all rigid transformations in $\mathrm{SE}(2)$ plus reflections.
% $ \mathrm{E}(2) = \mathbb{R}^2 \rtimes \mathrm{O}(2) $ ($\mathrm{O}(2)$ includes both rotations and reflections). 
This group allows orientation flips while preserving distances and angles.

% \vspace{1mm}
% \noindent
\vspace{2mm}
\noindent
\textbf{\textit{Similarity}} $[W_{\text{sim}} \coloneq \mathrm{Sim}^{+}(2)]$:
The similarity group includes all combinations of translations, rotations, and uniform scale.
This group preserves angles and orientation, but allows uniform changes of size.
% The orientation-preserving similarity group $ \operatorname{Sim}^{+}(2) = \mathbb{R}^2 \rtimes (\mathrm{SO}(2) \times \mathbb{R}_{+}) $. Transformations have the form $ g(x) = s R_{\theta} x + t $ with $ s > 0 $. 

\vspace{2mm}
\noindent
\textbf{\textit{Similarity + reflections}} $[W_{\text{sim-ref}} \coloneq \mathrm{Sim}(2)]$: The similarity plus reflections group includes all similarity transformations $\mathrm{Sim}^{+}(2)$ plus reflections.
% = \mathbb{R}^2 \rtimes (\mathrm{O}(2) \times \mathbb{R}_{+}) $.
This group allows orientation flips while preserving angles.

\vspace{2mm}
\noindent
\textbf{\textit{Affine}} $[W_{\text{aff}} \coloneq \mathrm{Aff}(2)]$:
The full affine group including all combinations of translation, rotation, non-uniform scaling, and shear.
% $= \mathbb{R}^2 \rtimes \mathrm{GL}(2, \mathbb{R}) $.
% Elements are of the form $ g(x) = A x + t $, where $ A $ is any invertible matrix.
This group preserves parallel lines and ratios along parallel lines.

\vspace{2mm}
\noindent
\textbf{\textit{Anisotropic similarity}} $[W_{\text{sim-ani}}]$: 
The group of all combinations of translation, rotation and non-uniform scale~\cite{steger2012least}.
This group is equivalent to the affine group but without shear. 
Note that trajectories are specified in the screen space coordinate frame. But many shapes (e.g. rectangles, right triangles, etc.) have canonical coordinate frames aligned with higher-level perceptual features such as right angles, symmetry axes, etc. 
The anisotropic similarity group allows non-uniform scale before or after rotation.
However, since we do not have access to the canonical orientation of a prototype trajectory, such non-uniform scales can appear as a shear \camready{(Supplemental Figure~\sref{fig:anisotropic}).}

\paragraph{Hierarchy of geometric warps.}
%\maneesh{Prev. sentences are out of place. We can reference the sable paper and ideas about how people talk about trajectories in the introduction and/or possibly related work. Here we just want to the mathematical setup/background.}
%\kenny{Need to add motivation of geometric warps}
The Lie transformation groups form a partially ordered hierarchy (Figure~\ref{fig:hierarchy}).
%We describe each group and then  
%organized by the relationships of their underlying transformation groups (Lie groups)~\cite{kaji2016concise, eade2014lie}, which we later use as inductive biases to recover shape families. \maneesh{we should not talk about how we are later going to use these things here. This section just needs to present mathematical machinery mostly independent from how we are going to use it. }
%We describe each warping function in detail below. \maneesh{I think there is some confusion here about whether the space is a single warping function or a space of warping functions. I think it is a space. Equivalently each space is a family that allows/admits a set of warps.}
%% ending
where the parent group of warps $W_i$ contains its child group of warps $W_j$, such that $W_j \subseteq W_i$. Thus, each child group is more restrictive than its parent group.
%This hierarchy forms a partially ordered set where $W_i \subseteq W_j$ if every transformation in $W_i$ is also in $W_j$.
%The  captures cases where warps at the same level are not subgroups of each other (Figure~\ref{fig:hierarchy}).
%
Additional Lie groups can be added to this hierarchy based on their subgroup relationships. For example, equi-affine transformations (area preserving warps $\mathrm{SA}(2) \subseteq \mathrm{Aff}(2)$) or projective transformations (collinearity preserving warps, $\mathrm{PGL}(2) \supseteq \mathrm{Aff}(2)$) could be added to the hierarchy. 

%
% This hierarchy reflects true subgroup relationships:
% $
% SE(2) = E^{+}(2) \triangleleft E(2) \triangleleft \operatorname{Sim}(2) \triangleleft W_{\text{anis-sim}} \triangleleft \mathrm{Aff}(2)
% $
% where each subgroup adds/removes transformation types, and each element of a subgroup is also an element of all supersets above it. Notably, for shapes where principal axes are well-defined, the anisotropic similarity group (rotation + axis-aligned non-uniform scaling + translation) provides greater modeling flexibility than similarity warps but less than general affine.

% The hierarchy is justified by the formal group inclusion $ G \subseteq H \iff \forall g \in G,\; g \in H $, with strict containment at each level. It reflects the geometric invariants preserved by each group, and is readily extended to projective ($ \mathrm{PGL}(3,\mathbb{R}) $), equi-affine, and more specialized warps as applications demand.

\paragraph{Warp-invariant distance metrics.}
For each transformation group $W$, we define a corresponding \textit{distance metric} $d_W$ that measures the distance between two trajectories while remaining invariant to warps in $W$. Specifically, given two trajectories $t_1$ and $t_2$, the $W$-invariant distance is:
\begin{equation}
d_W(t_1, t_2) = \min_{w \in W} \frac{1}{n}\sum_{i=1}^{n} \|w(t_{1,i}) - t_{2,i}\|^2
\end{equation}
where $t_{1, i}$ and $t_{2, i}$ denote the $i$-th points on trajectory $t_1$ and $t_2$ respectively.
We optimize over all warps $w \in W$ to find the best alignment between the transformed $t_1$ and $t_2$. This distance equals zero if and only if $t_2 \in \mathcal{F}(t_1, W)$. We implement our distance metrics using a generalized variant of the iterative closest point (ICP) algorithm~\cite{besl1992method,chen1992object,du2010affine}.
\camready{We represent trajectories in a 400 px $\times$ 400 px SVG and compute $d_W$ by resampling each trajectory to $n = 100$ by arc length.
Computing $d_W$ for one pair takes on average 67 milliseconds on a standard desktop CPU.}
In practice, due to discretization and alignment inaccuracies, we consider $t_2 \in \mathcal{F}(t_1, W)$ when $d_W(t_1, t_2)$ is less than a consistency threshold $\tau$. See Supplemental Section~\sref{sec:supp_method} for details.

%% file: sec/method.tex
\section{Method}
\label{sec:method}
% \maneesh{Need shorter section title}

With the definitions from the previous section, we have the machinery to apply self-consistency to the domain of motion trajectories.
Our procedure takes an input prompt that specifies a desired motion trajectory described as shapes.
In this work we assume that we can model the desired trajectory
as a shape family (Equation~\ref{eq:shape_fam}) using a transformation group from our hierarchy (Figure\,\ref{fig:hierarchy}). 
We extend ideas from self-consistency to automatically recover this shape family in an unsupervised fashion.
Access to this shape family allows us to produce better LLM responses (by selecting a member from the most consistently generated shape family) and verify whether other trajectories match the specification of the prompt (by checking membership in the family).

In the rest of this section, we present the components of our approach.
We first describe how we use an LLM to sample a collection of motion trajectories (Section~\ref{sec:met_sample}).
Next, we show that 
if we are given an oracle that can provide the \textit{correct} transformation group for the desired shape family, we can use self-consistency to identify a prototype trajectory from the collection of LLM-generated samples (Section~\ref{sec:met_proto}).
Then we consider the 
case when the \textit{correct} transformation group for the desired shape family is unknown, and show how we can estimate the transformation group using two different decision criteria (Section~\ref{sec:met_warp}).
Finally, in Section~\ref{sec:met_membership}, we describe how to verify whether a query trajectory is a member of a given shape family.

\subsection{Sampling Motion Trajectories with an LLM}
\label{sec:met_sample}

% As input, our procedure assumes a user has provided a natural language description that specifies a distribution of motion trajectories forming a well-defined shape family.
% \maneesh{should say this is well-defined in terms of our hierarchy. Actually should say this upfront -- see note above.}
Our first step uses an LLM to generate $N$ samples of motion trajectories conditioned on the input prompt (Figure~\ref{fig:teaser} right).  
%\maneesh{Fig 1 should explain  that given the prompt we generate multiple samples with an LLM and then cluster them ... etc.}
%Following\,\cite{ma2025mover}, 
We frame this step as a program synthesis problem: requesting the LLM to generate a motion graphics animation program, which uses functions from an API of high-level animation operations\,\cite{gsap}. 
Given a static SVG scene (e.g. a single element, centered on the canvas), we can then execute this program to produce a motion trajectory.

To encourage this collection of trajectory samples to exhibit representative modes of variations for the specified trajectory, we employ a prompting strategy that encourages diversity~\cite{zhang2025verbalized}.
Instead of repeatedly asking the LLM to produce \textit{independent} animation programs, we ask the LLM to produce $k < N$ programs in each response, and additionally specify that these responses should well-cover the ``tails'' of the correct distribution.
We repeat this sampling process in batches of size $k$ until the LLM has produced $N$ trajectories.
See Supplemental Section~\sref{sec:supp_gen_llm} for the full wording of the diversity sampling prompt.
%We provide full prompting details and further analysis in the supplemental material.

As discussed in Section~\ref{sec:preliminary}, for shape families defined with $W_{\text{sim-ani}}$ (anisotropic similarity) the orientation of the prototype trajectories is important. More specifically, if the prototype trajectory is oriented so that its canonical orientation is not aligned with the screen space x- and y-axes, the transformation group $W_{\text{sim-ani}}$ may appear to shear the trajectory even though shears are not included in the transformation group.
%This is because non-uniform scale along non-principal axes would form trajectories that fall outside of the shape family.
Therefore, for these families, we make an additional assumption that the LLM would generate at least some samples with their canonical orientations aligned with our screen-space coordinate frame.
%entire shape family by applying a fixed order of non-uniform scaling first followed by a rotation~\cite{steger2012least}.
%\maneesh{May need to talk about axis-aligned samples.}

% \paragraph{ground truth distance metric}
\subsection{Self-Consistent Trajectory Clustering}
\label{sec:met_proto}

If we have access to the \textit{correct} geometric transformation group for the desired shape family (i.e. $W$ is provided by some oracle), we describe how to use the collection of LLM sampled motion trajectories to choose a prototype.
Using the distance metric $d_W$ (associated with the transformation group $W$), we form a distance matrix $D$ by computing pairwise distances under $d_W$ between each pair of trajectories.
We send this distance matrix to the DBSCAN algorithm\,\cite{schubert2017dbscan} to produce clusters of consistent trajectories under $W$, setting the distance threshold as $\tau$. 
The cluster that has the most members corresponds to the set of trajectories that are most self-consistent with one another under $W$; producing a self-consistent generation is then equivalent to returning any member from this cluster.
To create a shape family that best aligns with the trajectories in this cluster under $W$, we choose the centroid sample trajectory (minimum average distance to all other cluster members under $d_W$) as the prototype trajectory $o$.  

\subsection{Decision Criteria for Selecting $W$}
\label{sec:met_warp}

Without any supervision or external knowledge, we do not have access to the correct warp function $W$ for a given prompt. 
Under different assumptions about the distribution of motion trajectories produced by the LLM, we can design decision criteria that will use our hierarchy of geometric transformation groups to determine an appropriate $W$.
We consider two such decision criteria: \DCmajor~and~ \DChier.
%In the rest of this section we describe two such decision criteria: \DCmajor~and~ \DChier.

% \vspace{1em}

\subsubsection{\DCmajor}
This procedure identifies the most restrictive transformation group for which a majority cluster first appears.
We order the transformation groups from most to least restrictive (ascending the hierarchy). 
Iteratively, for each group $W$, we cluster the LLM-sampled trajectories and record the size of the largest cluster. 
When we observe this clustering produces a dominant cluster that contains a strict majority of all samples (exceeding 50\%), we select the corresponding transformation group. 
This criterion finds the correct transformation group when:
(i) the LLM produces correct trajectories more frequently than incorrect ones;
(ii) the collection of sampled trajectories diversely covers the modes of variation possible within the transformation group. 
Under these assumptions, a majority cluster can arise only once the warp aligns with the true geometric structure of the shape family.

% \vspace{1em}

\subsubsection{\DChier}
In practice, assumption (ii) might not hold. So we offer an alternative criterion that better accounts for cases where the LLM samples do not cover the modes of variation within the true transformation group well.
We order the warp functions from least to most restrictive (descending the hierarchy).
Starting from the most permissive warp, we cluster the sampled trajectories, identify the largest cluster, and treat its members as example trajectories from the true shape family.
Then, progressively considering more restrictive warps, we check whether clustering would remove any members from the previously observed largest cluster.
We select the most restrictive warp for which the largest cluster remains intact.
This procedure introduces a new assumption: incorrect LLM generated trajectories will not cluster with correct ones under any warp in our hierarchy.
When this holds, the largest cluster begins to lose members precisely when the warp becomes overly restrictive for the true geometric structure of the underlying shape family.

\subsection{Shape Family Membership Verification}
\label{sec:met_membership}
% A prompt $p$ describing a type of mathematically well-defined shape (e.g., ``move in an elliptical orbit'') implicitly specifies a shape family $\mathcal{F}_p(o_p, W_p)$ with a a prototype trajectory $o_p \in \mathcal{F}_p$ as any member of that family and a warp type $W_p$ defining allowable geometric variations.

We can use the machinery described in the previous sections to solve verification problems.
For verification, we are given an input prompt that specifies a desired trajectory and shape family and a query motion trajectory \textit{t}. The goal of verification is to determine if \textit{t} matches the prompt. 
We start by applying our self-consistency approach as described above, to recover 
a shape family $\mathcal{F}$ from the input prompt, identifying both a transformation group $W$ and a prototype trajectory $o$ from a collection of LLM-generated samples. 
Then, given a query trajectory $t$, we solve the verification task by checking whether $t$ is a member of the shape family.
Formally, we say that $t$ is a valid response with respect to the input prompt if and only if $t \in \mathcal{F}(o, W)$.
Checking membership is done using the associated distance metric 
$d_W$: we compare the distance between $o$ and $t$, and say that $t$ is consistent with $\mathcal{F}$ if this distance is less than $\tau$.

%% file: sec/experiments.tex
\begin{figure*}[t]
    \centering
    \includegraphics[width=\linewidth]{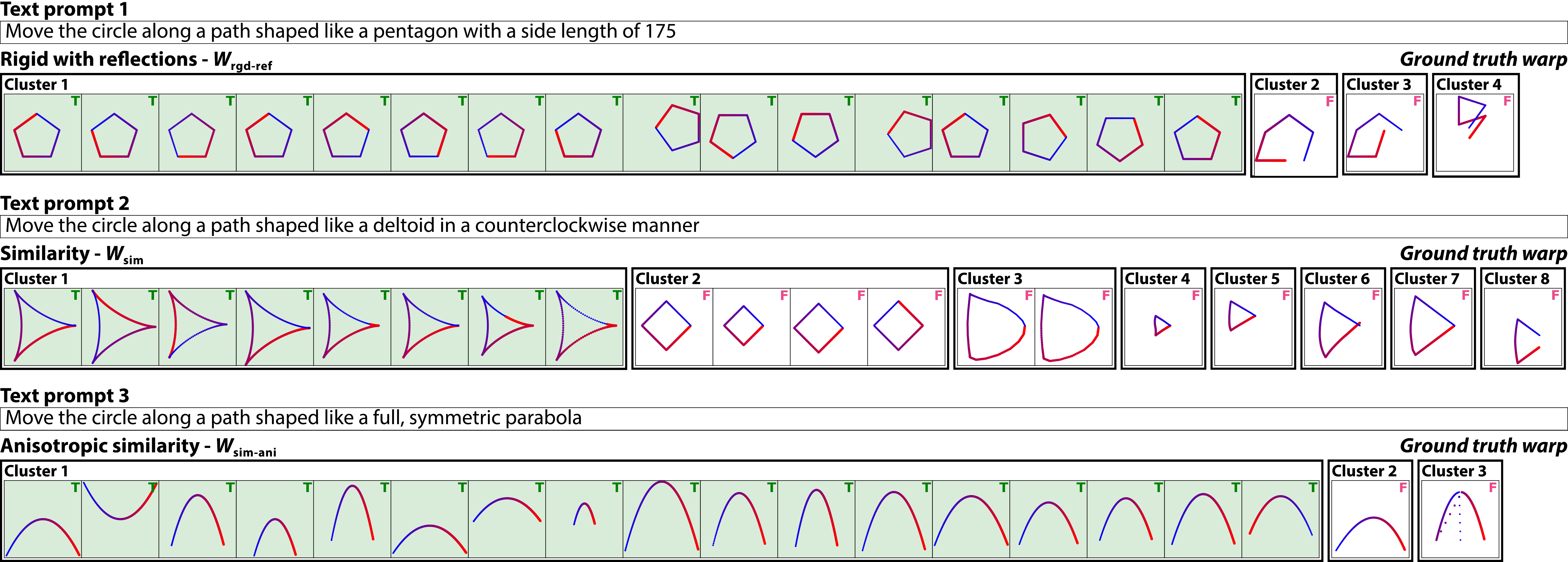}
    \caption{We present clustering results on LLM-generated samples from three example prompts in our dataset. Each trajectory has a ground truth label on the upper right. Clusters colored green are the chosen largest clusters, and we note that their sizes vary, sometimes less than half of the total number of trajectories (middle). 
    \camready{In the pentagon case, failed samples (Cluster 2--4) appear visually distinct from the true ones, while some of the failed deltoids (Cluster 4--8) look closer to the correct one with their triangular forms.}
    The rightmost parabola is a skewed one that would have been grouped into the largest cluster under the affine warp.}
    \label{fig:results}
\end{figure*}

\section{Results}
\label{sec:experiments}

To evaluate our extension of self-consistency for the domain of motion graphics, we develop a benchmark of prompts associated with ground-truth shape families (Section~\ref{sec:res_benchmark}).
We use this benchmark to validate that our approach provides performance boosts over baseline alternatives for both generation (Section~\ref{sec:res_gen}) and verification (Section~\ref{sec:res_verif}) settings.

\subsection{Constructing a Motion Trajectory Benchmark}
\label{sec:res_benchmark}

Comparing the performance of methods that generate or verify motion graphics trajectories requires a dataset that contains text prompts describing motion trajectories and their corresponding ground truth shape families.
As no such dataset is available, we synthetically design one using a template-based approach similar to MoVer~\cite{ma2025mover} and CLEVR~\cite{johnson2017clevr}.
% We provide visualizations for the prompts in our benchmark in the supplemental material.
\camready{% An expert manually verified our benchmark.
We provide the full details of our template-based construction process and visualize all 224 pairings in the supplemental material (Section~\sref{sec:supp_data}).}

\subsubsection{Prompts with ground truth shape families}
Our benchmark contains 224 prompts paired with ground-truth shape families.
To construct this benchmark, we first identified a set of \camready{35} geometric base shapes that people commonly use to describe the trajectories found in motion graphics animations, as documented by supporting literature~\cite{sable2022language}.
We then manually paired natural language descriptions of these geometric base shapes with a ground truth shape family (Eq.~\ref{eq:shape_fam}) formed by a prototype trajectory and a corresponding transformation group from our hierarchy. 
We were then able to procedurally create variations of these base shapes by appending the natural language descriptions with additional specifications that induce a change to the corresponding shape family through the transformation group. 
As an example, a prompt \textit{``Animate the blue circle to move along a path shaped like a circle''} corresponds with a shape family that uses ground truth warps from $W_{\text{sim-ref}}$ (similarity + reflections).
Modifying this prompt with additional specifications, \textit{``Animate the blue circle to move along a path shaped like a circle with a radius of 50 px''}, constrains the shape family so that it instead can only use warps from $W_{\text{rgd-ref}}$ (rigid + reflections).

\subsubsection{Test set of trajectories for verification}
To evaluate performance on the task of verifying motion trajectories, for each prompt we require query trajectories and associated labels for whether they match or mismatch the prompt.
To this end, we curated a test set of 2240 trajectories (10 per prompt).
For a given (prompt, query trajectory) pair, we used the ground-truth shape family corresponding with the prompt to provide an automatic label through membership checks (Section~\ref{sec:met_membership}).
To produce this set of query trajectories, we iteratively sampled an LLM (GPT-4.1 \& GPT-5) to produce motion trajectories conditioned on the input prompt (Section~\ref{sec:met_sample}).
We employed a rejection sampling-based approach until we found five true trajectories and five false trajectories for each prompt. 
%In the supplemental material we analyze how well human perception agrees with our template-based approach for automatically labeling these test-set trajectories. \maneesh{what analysis are we planning to do here?} \kenny{Something to confirm the 'correctness' of the dataset -- maybe have us internally rate alignment between prompts and }

% To evaluate if this set of trajectories are correctly labeled by our ground truth families, we randomly sampled one trajectory labeled as true and one labeled as false for each of the 224 prompts and asked human annotators to check if the trajectories correspond to the prompts. 
% \kenny{Reframe as in supplemental we analyze human agreement with our automatic procedure for defining shape families.}
% Please refer to the supplemental materials for more details.

\subsection{Motion Trajectory Generation}
\label{sec:res_gen}

\input{sec/table_synthesis}

We use our benchmark to measure the performance on the task of our self-consistency-based motion trajectory generation methods. 
We consider a baseline generation method that uses an LLM to sample a single motion trajectory for each prompt in our evaluation set (LLM direct).
We compare this baseline against our approach that samples an LLM multiple times (Section~\ref{sec:met_sample}) to identify the most self-consistent trajectories (Section~\ref{sec:met_proto}), setting the number of samples to $N=19$.
We evaluate the performance of these methods with an accuracy metric, by checking whether the generated trajectory matches the prompt or mismatches the prompt using a membership check against the corresponding ground-truth shape family.
%\maneesh{Not sure I follow prev. sentence. Which method is it talking about?} \kenny{This is supposed to indicate general method for task of verification} \maneesh{Let's quickly chat about it on zoom.}
%One approach is to use an LLM to produce a single motion trajectory, we call this strategy LLM-Direct.
%We compare this baseline against our approach that samples an LLM multiple times (Section~\ref{sec:met_sample}) to identify the most self-consistent trajectories (Section~\ref{sec:met_proto}), setting $N=19$.

We report the results of this experiment in Table~\ref{tab:accuracy}, where we split the analysis by the LLM used for sampling (GPT-4.1 vs GPT-5).
Trajectories generated directly by an LLM without self-consistency (top-row) produce accuracy rates of 62.1\% (GPT-4.1) and 79.1\% (GPT-5).
Our self-consistency methods improve the generation accuracy for a given sampling LLM.
In the oracle setting, where self-consistency is computed under the ground-truth transformation group $W$, we observe improvements of 6.4 and 4.4 percentage points for GPT-4.1 and GPT-5 respectively.
In the unsupervised setting, when we use the different decision criteria to estimate $W$ (Section~\ref{sec:met_warp}) we observe these criteria still provide substantial boosts. 
\DCmajor~closely matches the oracle setting, achieving 5.9 and 4.2 percentage point improvements for GPT-4.1 and GPT-5 respectively. 
\DChier~performs slightly worse, but still significant, with 4.6 and 3.5 percentage point improvements for GPT-4.1 and GPT-5 respectively. 
%\maneesh{Say something here or at end of qualitative paragraph explaining what kinds of errors our approach is making.}

% \kenny{paragraph to analyze qualitative trends TBD}
Figure~\ref{fig:results} shows qualitative results under the oracle geometric warps.
In the pentagon case, as the false trajectories are distinct in a way that no geometric warps can map them to the correct ones and the largest cluster forms an absolute majority, both \DCmajor{} and \DChier{} would correctly choose the ground truth warp of rigid with reflections.
For the deltoids, although the samples in the bottom row have a triangular form, no geometric warps can further merge any clusters together.
As a result, both \DCmajor{} and \DChier{} would select the right warp, even when the largest cluster is less than half of the total number of trajectories.
The parabola case is where \DCmajor{} would succeed, as the anisotropic similarity warp is the most restrictive one to produce a largest cluster containing at least half of the trajectories.
However, the bottom left trajectory is a skewed parabola and can be merged into the largest group under affine, causing \DChier{} to fail.

\subsection{Motion Trajectory Verification}
\label{sec:res_verif}

% task: verify if a trajectory follows a given prompt. the test trajectories are from the curated test set.
\input{sec/table_verification}

We can also use our benchmark to measure the performance on the task of motion trajectory verification.
In this setting, a method is presented with a prompt and a query motion trajectory, and is tasked with determining whether the trajectory matches the prompt.
We consider two VLMs (GPT-4.1 and GPT-5) as baseline verification methods. 
We can feed a VLM with a prompt and a static rendering of a motion trajectory generated from that prompt, and request that the VLM should indicate whether these two items match or not (see Supplemental Section~\sref{sec:supp_verify_vlm} for details).
We compare these baselines against our self-consistency-based method for verification, as described in Section~\ref{sec:met_membership}.
We use GPT-5 as the LLM sampling model, and once again set $N=19$.

We compute precision, recall, and F1 scores by comparing labels predicted by each method against the ground truth labels in our benchmark, and report the results of this experiment in Table~\ref{tab:results}.
We observe that using GPT-4.1 as a verifier is poorly calibrated: the model’s predicted prevalence (the rate at which it predicts True) is 90\%, 
%\maneesh{what is prevalence?} \kenny{the rate at which it predicts True}
far above the true base rate of the data at 50\%. This results in a high recall (96.9), at the expense of a low precision (62.0).
GPT-5 produces more balanced precision (74.0) and recall (84.7), achieving an F1 score of 79.
With oracle access to the ground-truth $W$, our self-consistency approach for verification is able to achieve better performance (F1 of 85.6).

\subsection{Discussion}

\camready{
We discuss how to choose a decision criterion and set the hyperparameters for our approach.
See the supplemental material (Sections~\sref{sec:supp_dc_failure} and \sref{sec:supp_dbscan}) for more details.

\paragraph{Choosing decision criteria.}
Our extension of self-consistency works in an unsupervised setting, where the correct transformation group $W$ is not known a priori.
% In Section~\ref{sec:met_warp}, we introduce two decision criteria to estimate $W$, but which criteria should be used for a particular task?
We allow users to flexibly choose between the two decision criteria for $W$ estimation introduced in Section~\ref{sec:met_warp} that would best support their downstream application needs.
One should use \DCmajor{} to prioritize precision, and use \DChier{} to balance improved recall while maintaining reasonable precision.
The design of \DCmajor~leads it to have a bias for selecting overly restrictive transformation groups, as it traverses our hierarchy of warps in an ascending order.
For verification (Table~\ref{tab:results}), we find that this form of conservatism leads to the best precision (85.8) at the expense of recall (66.1).
Conversely, \DChier~traverses our hierarchy of warps in a descending order, which finds a nice trade-off between precision and recall, allowing it to achieve the best F1 score for the unsupervised verification task (84.6).
These trends are further supported by an analysis of failure modes of these decision criteria: when they choose an incorrect $W$, \DCmajor{} selects an overly restrictive group 95.6\% of the time while \DChier{} selects an overly permissive one 80.6\% of the time.

%Our extension of self-consistency works in an unsupervised setting, where the correct transformation group $W$ is not known a priori.
%The two decision criteria we propose to estimate $W$ have different properties in terms of precision and recall to support different downstream application needs (Table~\ref{tab:accuracy} and \ref{tab:results}).
%%
%% Users can choose decision criteria to support different downstream application needs.
%One should use \DCmajor{} to prioritize precision, such as when generating a correct trajectory or verification with high true positive rate, and use \DChier{} to prioritize recall while maintaining reasonable precision.
%We support this by an analysis of failure modes of these decision criteria: when they err, \DCmajor{} selects an overly restrictive group 95.6\% of the time while \DChier{} selects an overly permissive one 80.6\% of the time. See the supplemental material (Section~\sref{sec:supp_dc_failure}) for the full breakdown.

\begin{figure}[t]
    \centering
    \includegraphics[width=\linewidth]{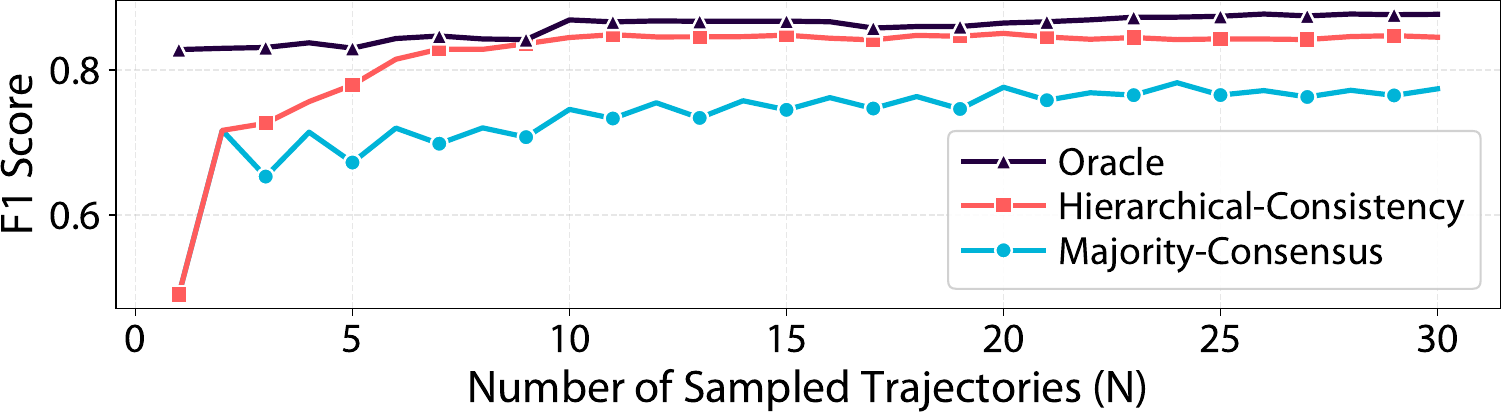}
    \vspace{-6.5mm}
    \caption{F1 scores across different numbers of sampled trajectories $N$. Decision criteria performances stabilize after $N=10$.}
    \label{fig:num_trajectories}
    \vspace{-5mm}
\end{figure}

\paragraph{Choosing hyperparameters.}
Our approach has two main hyperparameters, the number of sampled trajectories $N$ and the clustering threshold $\tau$.
For $N$, our algorithm works well with as few as $N=10$ trajectories.
% , and sampling more continues to improve performance, although only marginally.
We repeat the verification experiment of Section~\ref{sec:res_verif} while varying $N$ from 1 to 30, using GPT-5 as the sampling LLM.
As shown in Figure~\ref{fig:num_trajectories}, at low $N$, F1 increases sharply as additional samples give the clustering more signal; performance then stabilizes around $N=10$, with only marginal gains beyond.
This trend holds across all ways of choosing a warp, showing that a modest sample budget of 10 is typically sufficient.
While we perform DBSCAN clustering with a small distance threshold $\tau$ to account for discretization errors, our approach is not sensitive to the choice of its value.
To show this, we repeat the verification experiment with \DChier{} while sweeping $\tau$ from 0.25 to 8.0, a 32$\times$ range.
We observe that F1 drops by only 7.2 percentage points across the full sweep.}
% \kenny{I think this sentence should go higher}}

%% file: sec/table_synthesis.tex
\begin{table}[t]
\centering
% \small
\caption{Motion trajectory generation experiment (see Sec.~\ref{sec:res_gen}). We evaluate the accuracy of LLM-generated motion trajectories (GPT-4.1 and GPT-5) under different decoding strategies. Under different decision criteria for selecting a transformation group $W$, our self-consistency approach improves upon the baseline alternative of directly generating a single sample (LLM-Direct).
%Accuracy comparison of our self-consistency methods with different decision criteria to select the warp $W$ on trajectories generated by GPT-4.1 and GPT-5.
%Accuracy is computed as the average of percentages of the number of true trajectories in a chosen cluster across prompts (for LLM-Direct, the ``chosen cluster'' includes all trajectories for a prompt.
}
\label{tab:accuracy}
\vspace{-2mm}
\resizebox{\columnwidth}{!}{%
\begin{tabular}{llcc}
\toprule
Method & Decision Criteria & GPT-4.1 & GPT-5 \\
\midrule
LLM-Direct  & --- & 62.1 & 79.1 \\
Ours & \DCmajor{} & \textbf{68.0} & \textbf{83.3} \\
Ours & \DChier{} & 66.7 & 82.6 \\
\midrule
Ours & Oracle  & 68.5 & 83.5 \\
\bottomrule
\end{tabular}%
}
\vspace{-3mm}
\end{table}

% \begin{tabular}{lc}
% \toprule
% Method & Accuracy \\
% \midrule
% LLM-Direct & 79.1\% \\
% GT Distance Metric & \textbf{83.5\%} \\ 
% Simple Majority & 83.3\% \\
% Hierarchical & 82.6\% \\
% \bottomrule
% \end{tabular}

%% file: sec/table_verification.tex
\begin{table}[t!]
\centering
% \footnotesize
% \small
% \scriptsize
\caption{
Motion trajectory verification experiment (see Sec.~\ref{sec:res_verif}).
We compare our self-consistency-based approach for verifying whether a trajectory matches an input prompt against the alternative of using a VLM (GPT-4.1 or GPT-5).
%\kenny{TODO}Performance comparison of our self-consistency methods with different decision criteria for $W$. The metrics are computed against ground truth verification labels produced by comparing against ground truth shape families of each prompt in our dataset.
% \kenny{A proposal for how to re-organize this table}
}
\label{tab:results}
\vspace{-2mm}
\resizebox{\columnwidth}{!}{%
\begin{tabular}{llccc}
\toprule
Method & Decision Criteria &  Precision & Recall & F1 Score \\
\midrule
GPT-4.1  &--- & 62.0 & \textbf{96.9} & 75.6 \\
GPT-5  & --- & 74.0 & 84.7 & 79.0 \\
Ours & \DCmajor{} & \textbf{85.8 }& 66.1 & 74.6 \\
Ours & \DChier{} & 80.5 & 89.0 & \textbf{84.6} \\
\midrule
Ours & Oracle  & 87.9 & 83.3 & 85.6 \\
\bottomrule
\end{tabular}%
}
\vspace{-3mm}
\end{table}

% \begin{table}[t]
% \centering
% \caption{Performance comparison of different methods on the verification task.}
% \label{tab:results}
% \begin{tabular}{lccc}
% \toprule
% Method & Precision & Recall & F1 Score \\
% \midrule
% VLM Baseline GPT-5 & 0.740 & 0.847 & 0.790 \\
% VLM Baseline GPT-4.1 & 0.620 & \textbf{0.969} & 0.756 \\
% GT Distance Metric & \textbf{0.879} & 0.833 & \textbf{0.856} \\
% Simple Majority & 0.858 & 0.661 & 0.746 \\
% Hierarchical & 0.805 & 0.890 & 0.846 \\
% \bottomrule
% \end{tabular}
% \end{table}

%% file: sec/discussion.tex
% \section{Discussion}
% \label{sec:discussion}

% \vspace{2em}

\section{Conclusion}
\label{sec:conclusion}

We introduced an approach that extends self-consistency to the visual domain of motion graphics animations.
The core challenge is determining when motion trajectories should be considered consistent with one another.
We propose a shape-specific notion of consistency by identifying a hierarchy of Lie transformation groups that can be used to define common geometric shape families.
Each group is paired with a distance metric that is invariant under its warps, allowing us to cluster LLM-produced motion trajectories in a self-consistent fashion.
We further proposed multiple decision criteria that automatically select an appropriate transformation group for a given prompt by analyzing the behavior of these clusters.
We empirically observed that our training-free, unsupervised extension of self-consistency improves performance on both motion trajectory generation and verification tasks over LLM and VLM baselines.

\paragraph{Limitations.}
% limitations on assumptions for decision criteria
Like prior work on self-consistency~\cite{wang2022self}, our approach cannot recover from all mistakes made by the sampling LLM. 
As discussed in Section~\ref{sec:met_warp}, our procedure may fail when certain assumptions on the distribution of LLM-generated trajectories are not met.
We analyze these failure modes in the supplemental material (Section~\sref{sec:supp_lim}).

% limitation in shape families
Our approach handles prompts describing shape families characterized by a single prototype and warp, such as ``move in a circular orbit'' (circle under similarity warp) and ``trace an Archimedean spiral'' (Archimedean spiral under similarity warp with reflections).
Ambiguous descriptions (e.g., ``move in a curved path'') or prompts with multiple disjoint families are outside our scope. 
For example, the prompt ``move in a heptagram,'' ``heptagram'' maps to both heptagram \{7/2\} and heptagram \{7/3\} and no geometric warps can bring one to the other. 
\camready{When multiple prototypes exist, each prototype's cluster is smaller. This makes it harder for our decision criteria to distinguish within-family clusters from outlier ones.
We experimented with a simple modification to \DChier{} that returns all largest clusters whose sizes are within 20\% of one another, improving F1 from 71.0 to 88.9 on multi-prototype prompts. See the full analysis in Supplemental Section~\sref{sec:multi_prototype_experiment}.}

% \maneesh{Ideally a figure showing this limitation of the heptagram should be included as most readers probably won't know what the two forms really area. Ideally in supplemental materials we should show some examples of prompts that have more than 1 prototype.}

% future work 
\paragraph{Future work.}
Looking ahead, we view this work as a template for extending self-consistency beyond natural–language reasoning tasks, where consistency is typically defined narrowly through identity matching. 
Generalizing this paradigm requires identifying a domain-relevant notion of consistency, which likely needs to be flexible with respect to prompt-level semantics.
Our proposed solution for the domain of motion graphics leverages a hierarchy of common geometric transformation groups, which may be directly extendable for related spatial and shape-based domains.
Exploring how this general framework might adapt over a broader set of application areas is a promising direction.
Such efforts may prove especially impactful, as beyond better generation, the ability to extend consistency over distributions of related samples opens a path toward automatic, training-free verification.

%% file: sec/acknowledgments.tex
\section*{Acknowledgments}
\camready{Jiaju Ma was supported by the Stanford Graduate Fellowship.
% NSF FMITF grant and also Hoffman Yee grant and brown Institute
This work was partially supported by the Brown Institute for Media Innovation at Stanford University, NSF Awards \#2219864 and \#2211258, ONR YIP N00014-24-1-2117, AFOSR YIP FA9550-23-1-0127,
the Stanford HAI Hoffman-Yee Research Grant (Integrating Intelligence),
and the HAI-HPI Program on Artificial Intelligence (AI) and Human-Computer Interaction (HCI).}
% \jiaju{is this the right name of the hoffman yee grant?}

%% file: suppl_sec/benchmark_details.tex
\section{Motion Trajectory Benchmark Details}
\label{sec:supp_data}
% \maneesh{Should reiterate the assumptions of this benchmark up front. -- Excercise well defined shape families described by a transformation group and a single prototype. Might be useful to describe at end of section shape family examples that don't fit this assumption. Do we show what happens on a few such exampels later on -- I think we need to add these if we didn't show this in the main paper.}
%
Our benchmark consists of 224 motion trajectory prompts describing 35 different basic shapes, as shown in Figure~\ref{fig:benchmark_shapes}.
Each prompt is paired with a ground truth shape family following the definition of Equation 1 in the main paper -- one prototype trajectory and a transformation group.
\camready{
Table~\ref{tab:prompt_stats} reports the number of prompts associated with each transformation group.
The distribution is determined by the geometric properties of the base shapes (e.g., a circle cannot have anisotropic similarity or affine warps).
}
We discuss and show examples of applying our approach to shape families that do not follow this definition in Section~\ref{sec:family_multiple}.
We illustrate all 224 prompts with their ground truth shape families in our benchmark in Figure~\ref{fig:bench_mark_1}--\ref{fig:bench_mark_7}.

We use a template-based approach to generate the prompts, following MoVer~\cite{ma2025mover} and CLEVR~\cite{johnson2017clevr}.
Each prompt starts with the main sentence \texttt{Animate the <SVG element> to move along a path shaped like <shape name>.}
We manually pair a prompt to its ground truth shape family consisting of a Lie group of transformations and a prototype trajectory
based on the shape properties.
%\maneesh{need to use wording from the paper -- geometric transformation group or some such. I think the phrasing needs to be more like -- the prompt maps to ground truth a shape family consisting of the geometric transformation group and a prototype trajectory -- or some such. -- the prompt maps to a ground truth family. We specify the ground truth family by hand. } 
For example,  we pair ``circle'' with a circular trajectory and the warp $W_{\text{sim-ref}}$ (similarity with reflections), ``ellipse'' with an elliptical path and $W_{\text{sim-ani}}$ (anisotropic similarity), and ``parallelogram'' to a parallelogram path with $W_{\text{aff}}$ (affine).
We note that, for shapes that are periodic (sine and cosine waves) or self-repeating (Archimedean and logarithmic spirals), the prompts need to specify the number of periods or turns, as changes in these values effectively create different shape families (e.g., a spiral with 2 turns cannot be geometrically warped into a spiral with 3 turns).
We add a new sentence after the main sentence for this value: \texttt{Complete <periods/turns> of <shape name>.}

To introduce prompt variations that exercise all geometric warps in our hierarchy, we add modifiers to a prompt in the form of additional specifications on a shape.
% \maneesh{We are being too loose about the distinction between geometric transformation groups and shape families}
%
% \maneesh{There are some problems with plurals. Would be good to check the grammar for the entire document.}
The first type of specifications is \textit{shape orientation}, where we require a shape to be traversed in either a clockwise or counterclockwise manner.
Adding this specification removes reflections from the warps.
We use the following template sentence: \texttt{Traverse the path in a <orientation> manner.}

The second type of specifications is \textit{shape size}.
For some shapes, their sizes can be fully specified by one parameter (e.g., radius for circles and side length for polygons). 
Others require two parameters, such as width and height for rectangles and triangles, and horizontal extent and amplitude for sine and cosine waves.
Fully specifying a shape's size reduces the transformation group to either $W_{\text{rgd}}$ (rigid) or $W_{\text{rgd-ref}}$ (rigid with reflections).
Alternatively, we can specify a shape's size in terms of ratios (e.g., a rectangle with a 4:3 aspect ratio), and doing so changes the transformation group to either $W_{\text{sim}}$ or $W_{\text{sim-ref}}$.
We append a prepositional phrase to the main sentence when adding size specifications: \texttt{a <shape name> with <size param name> of <size param value>} (e.g., \textit{``a rectangle with a width of 50 px and a height of 80 px.''}).

\begin{table}[t]
\centering
\caption{\camready{Number of prompts per warp in our benchmark.}}
\label{tab:prompt_stats}
\vspace{-2mm}
\small
\begin{tabular*}{\columnwidth}{@{\extracolsep{\fill}}ccccccc@{}}
\toprule
$W_{\text{rgd-ref}}$ & $W_{\text{rgd}}$ & $W_{\text{sim}}$ & $W_{\text{sim-ref}}$ & $W_{\text{sim-ani}}$ & $W_{\text{aff}}$ & Total \\
\midrule
105 & 44 & 38 & 23 & 12 & 2 & 224 \\
\bottomrule
\end{tabular*}
\end{table}

% - shape family
%     - size 
%         - width, height, radius, side length, amplitude, x-extent
%         - ratio
%     - path orientation (clockwise, counterclockwise)
%     - period/turn

% \jiaju{- [high] figure for verification of benchmark (example prompt, prototype + warp, query trajectories with labels) }

%% file: suppl_sec/experiment_details.tex
%\section{Extended Experimental Analysis}
\section{Decision Criteria Alternatives}
\label{sec:supp_exp}

% \maneesh{Not sure why these are called ablations -- they just seem like other baseline decision criteria.}
% \maneesh{Stick to present tense throughout this document. Here change from past tense to present.}
% \kenny{Addressed}
In Section 4 of the main paper we describe how our method is able to recover a shape family from an input prompt. 
A critical component of this procedure is how to estimate which transformation group $W$, from our hierarchy of Lie transformation groups (Figure 3, main paper), is most appropriate.
We introduce two decision criteria for selecting $W$, based on underlying assumptions made in the sampled motion trajectory distributions produced by an LLM: \DCmajor~(Section 4.3.1) and \DChier~(Section 4.3.2).  
How do we know whether the decision criteria we propose select useful transformation groups? 
One way to justify our design of these procedures is by comparing their performance against very simple decision criteria alternatives.
To this end, we consider two additional `baseline' decision criteria, which we expect will recover worse shape families compared with \DCmajor~and~\DChier.
\textit{\DCmost} is a criterion that always selects the most restrictive transformation group in our hierarchy (rigid).
\textit{\DCleast} is a criterion that always selects the least restrictive transformation group in our hierarchy (affine).
% \maneesh{I don't follow the reasoning. How do these additional baselines justify anything. Need to explain what comparisons to these baselines will be expected to tell us. }
% \kenny{Took a pass at addressing these concerns}

% \maneesh{Need to add boldface in the tables.}
% \kenny{I added boldface in Table 1 -- I'm not sure we want boldface in Table 2, because of the trend in the precision column and recall column don't favor our method. We explain this in text, but if we bold alternative methods, we might give reviewers who only pass over the results a bad impression. }

\paragraph{Motion trajectory generation}

With the same experimental set-up as described in Section 5.2 of the main paper, we compare how the different decision criteria aid in the task of motion trajectory synthesis.
We report results of this experiment in Table~\ref{tab:accuracy_suppl}.
As a reminder, the metric we track is the ``accuracy'' of the generations, i.e., whether the production is a member of the ground-truth shape family from our benchmark dataset. 
For better numerical stability, we compute this accuracy as the ratio of true to false trajectories in the chosen cluster, averaged across tasks in the dataset. 
Note that for LLM-Direct, we say that the ``chosen cluster'' includes all trajectories produced from the same prompt.

We find that the two baseline decision criteria do strictly worse compared with the more sophisticated alternatives.
Even these simple ways of choosing $W$ do provide some benefit within the self-consistency paradigm in terms of improving the accuracy of the LLM's generations.
For instance, \DCmost~is able to increase the accuracy over LLM-Direct by between 3 and 4 absolute percentage points.
These results show that even with sub-optimal choices of $W$, self-consistency can benefit from the greater flexibility afforded in terms of checking whether generations match beyond identity comparisons. 

\paragraph{Motion trajectory verification}
Following the experimental set-up as described in Section 5.3 of the main paper, we compare how these alternative decision criteria perform on the task of motion trajectory verification. 
We report results of this experiment in Table~\ref{tab:results_suppl}.
\DChier~continues to achieve the best F1 score, finding a nice balance between precision and recall.
One way to view \DCleast~is as a partial implementation of \DChier; this simplification to never descend the hierarchy results in a steep decrease in precision (15 points) with only moderate increases to recall (3 points). 
In a similar fashion, \DCmost~can also be considered a partial implementation of \DCmajor. 
Always choosing the most restrictive transformation group provides a marginal benefit for precision (.3 points), but it dramatically drops the associated recall (24 points).

\input{suppl_sec/table_synthesis_suppl}

\input{suppl_sec/table_verification_suppl}

% \subsection{Ablation on Diversity Sampling}

% \jiaju{TODO: run sampling without diversity sampling}
% \kenny{Note: remove prior subsection heading if we only have one subsection}

% \subsection{Possible ablation studies:}
% % - [high] choosing distance metric (always most restrictive, always most lenient) for both synthesis and verification

% - [med] ablation study replacing diversity sampling with typical sampling

% - [low] number of sampled trajectories (N=?, k=?)

% - [low] report standard deviation for SC vs LLM

% - [no - high] perform verification with GPT 4.1 samples

% - [no - low] other LLMs? (beyond openAI)

% - [no - med/low] tau value on both synthesis / verification

%% file: suppl_sec/table_synthesis_suppl.tex
\begin{table}[t]
\centering
% \small
\caption{Motion trajectory generation experiment with two additional baseline decision criteria of always selecting the most or least restrictive warps. We observe that they perform strictly worse than the two methods proposed in the main paper.%
% (see Sec.~\ref{sec:res_gen}). We evaluate the accuracy of LLM produced motion trajectories (GPT-4.1 and GPT-5) under different decoding strategies (Methods). Under different ways of estimating a transformation group $W$, our self-consistency approach improves upon the baseline alternative of directly generating a single sample (LLM-Direct).
%Accuracy comparison of our self-consistency methods with different decision criteria to select the warp $W$ on trajectories generated by GPT-4.1 and GPT-5.
%Accuracy is computed as the average of percentages of the number of true trajectories in a chosen cluster across prompts (for LLM-Direct, the ``chosen cluster'' includes all trajectories for a prompt.
}
\label{tab:accuracy_suppl}
\resizebox{\columnwidth}{!}{%
\begin{tabular}{llcc}
\toprule
Method & Decision Criteria & GPT-4.1 & GPT-5 \\
\midrule
LLM-Direct  & --- & 62.1 & 79.1 \\
Ours & \DCmajor{} & \textbf{68.0} & \textbf{83.3} \\
Ours & \DCmost{} & 66.5 & 82.5 \\
Ours & \DChier{} & 66.7 & 82.6 \\
Ours & \DCleast{} & 64.6 & 82.1 \\
\midrule
Ours & Oracle  & 68.5 & 83.5 \\
\bottomrule
\end{tabular}%
}
\vspace{1cm}
\end{table}

% \begin{tabular}{lc}
% \toprule
% Method & Accuracy \\
% \midrule
% LLM-Direct & 79.1\% \\
% GT Distance Metric & \textbf{83.5\%} \\ 
% Simple Majority & 83.3\% \\
% Hierarchical & 82.6\% \\
% \bottomrule
% \end{tabular}

%% file: suppl_sec/table_verification_suppl.tex
\begin{table}[t!]
\centering
% \footnotesize
% \small
% \scriptsize
\caption{
Motion trajectory verification experiment (Table 2 of the main paper) with additional decision criteria of always selecting the most or least restrictive warps.%
% \jiaju{maybe add a sentence about what trend we see}
% (see Sec.~\ref{sec:res_verif}).
% We compare our self-consistency-based approach for verifying whether a trajectory matches an input prompt against the alternative of using a VLM (GPT-4.1 or GPT-5).
%\kenny{TODO}Performance comparison of our self-consistency methods with different decision criteria for $W$. The metrics are computed against ground truth verification labels produced by comparing against ground truth shape families of each prompt in our dataset.
% \kenny{A proposal for how to re-organize this table}
}
\label{tab:results_suppl}
\resizebox{\columnwidth}{!}{%
\begin{tabular}{llccc}
\toprule
Method & Decision Criteria &  Precision & Recall & F1 Score \\
\midrule
GPT-4.1  &--- & 62.0 & 96.9 & 75.6 \\
GPT-5  & --- & 74.0 & 84.7 & 79.0 \\
Ours & \DCmajor{} & 85.8 & 66.1 & 74.6 \\
Ours & \DCmost{} & 86.1 & 42.1 & 56.5 \\

Ours & \DChier{} & 80.5 & 89.0 & 84.6 \\
Ours & \DCleast{} & 65.0 & 91.8 & 76.1 \\
\midrule
Ours & Oracle  & 87.9 & 83.3 & 85.6 \\
\bottomrule
\end{tabular}%
}
\vspace{1cm}
\end{table}

% \begin{table}[t]
% \centering
% \caption{Performance comparison of different methods on the verification task.}
% \label{tab:results}
% \begin{tabular}{lccc}
% \toprule
% Method & Precision & Recall & F1 Score \\
% \midrule
% VLM Baseline GPT-5 & 0.740 & 0.847 & 0.790 \\
% VLM Baseline GPT-4.1 & 0.620 & \textbf{0.969} & 0.756 \\
% GT Distance Metric & \textbf{0.879} & 0.833 & \textbf{0.856} \\
% Simple Majority & 0.858 & 0.661 & 0.746 \\
% Hierarchical & 0.805 & 0.890 & 0.846 \\
% \bottomrule
% \end{tabular}
% \end{table}

%% file: suppl_sec/shape_family_limitation.tex
\begin{figure*}[th]
    \centering
    \includegraphics[width=\textwidth]{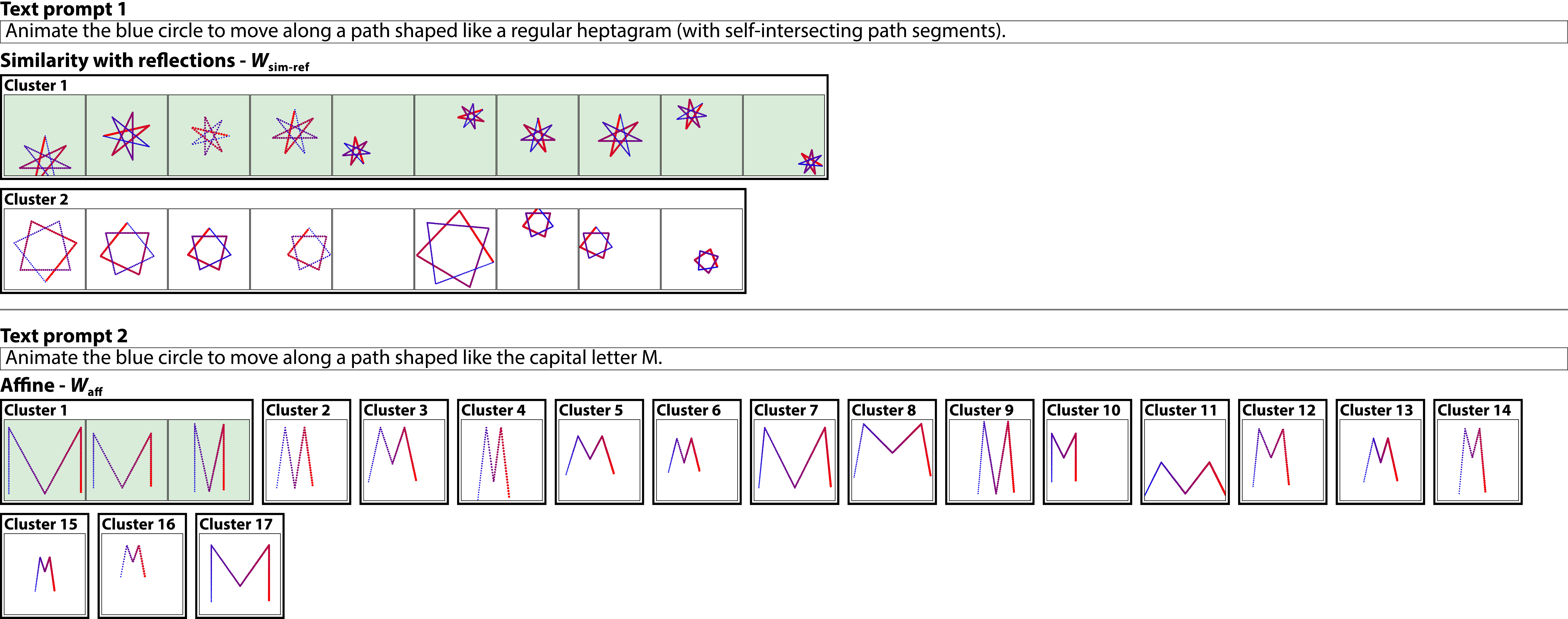}
    \caption{We demonstrate a limitation of our self-consistency approach in terms of the definition of a shape family.
    In text prompt 1, ``heptagram'' has two distinct prototypes: heptagram \{7/2\} and heptagram \{7/3\}.
    Under the warp $W_{\text{sim-ref}}$, cluster 1 contains correct \{7/3\} trajectories, while cluster 2 has \{7/2\}.
    While, for generation, our approach would produce a correct interpretation of heptagram, we would mark all heptagram \{7/2\} as false for verification.
    Similarly, for the letter Ms in text prompt 2, our approach would generate a correct form of the letter, but verify all other letter Ms that are not an affine warp of the shapes in cluster 1 as false.
    }
    \label{fig:limitations}
\end{figure*}

\section{Shape Family Limitations}
\subsection{Shape Families Requiring Multiple Prototypes}
\label{sec:family_multiple}
% prompts with multiple disjoint families are out- 591side our current scope. For example, the prompt“move in 592a heptagram,” “heptagram” maps to both heptagram {7/2} 593and heptagram {7/3} and no geometric warps can bring one 594to the other.

In our work, we define a shape family $\mathcal{F}(o,W)$ in Equation 1 of the main paper with one prototype and one geometric warp, and have demonstrated that a wide variety of common shape families (Figure~\ref{fig:benchmark_shapes}) that people use to describe motion trajectories falls under this definition~\cite{sable2022language}.

However, there are shape families that cannot be represented this way.
One type of such families requires multiple but finitely many prototypes under one particular warp.
For example, as described in the main paper, the shape family ``heptagram'' needs two distinct prototypes: heptagram {7/2} and heptagram {7/3} under the similarity warp (Figure~\ref{fig:limitations}).
% and no geometric warps can bring one to the other.
Another type of families either needs an infinite amount of prototypes under one warp, or one prototype with some sophisticated, non-geometric warp.
An example of this is the family of any letter of the English alphabet, like the capital letter M (Figure~\ref{fig:limitations}).
As demonstrated by Hofstadter and McGraw~\cite{hofstadter1993letter}, since there are countless ways to produce a trajectory that a human would perceive as an 'M-shape', we either need an infinite number of prototypes or a warp beyond the geometric transformations we consider.
% which one can write the letter M

In Figure~\ref{fig:limitations}, we prompted for the heptagram and capital M shapes.
In both cases, while the largest clusters do contain correct trajectories for generation, applying them for verification would result in limited recall, as correct shapes that cannot be mapped to by the single prototype and the warp would be labeled as false.

\subsubsection{Extension to Multi-Prototype Shape Families}
\label{sec:multi_prototype_experiment}

\camready{
For shape families requiring a finite number of prototypes under one warp,
% When multiple prototypes exist, each prototype's cluster is smaller, making it harder for our decision criteria to distinguish within-family clusters from outlier clusters.
we experimented with a simple modification to $\DChier{}$ that returns all largest clusters whose sizes are within 20\% of one another, instead of only the single largest cluster.
We tested this modification on four multi-prototype prompts under $W_{\text{sim-ref}}$:
heptagram (2 prototypes: \{7/2\}, \{7/3\}; Figure~\ref{fig:limitations}),
heptagram outline (2 prototypes: \{7/2\}, \{7/3\}),
octagram outline (2 prototypes: \{8/2\}, \{8/3\}),
and hendecagram (4 prototypes: \{11/2\}, \{11/3\}, \{11/4\}, \{11/5\}).
With regular $\DChier{}$, the verification F1 score on these multi-prototype prompts is 71.0.
After applying the 20\% tolerance modification, F1 improves to 88.9.
Meanwhile, the modification has a negligible effect on single-prototype prompts, with F1 decreasing only slightly from 84.6 to 84.5, as the tolerance occasionally causes an additional outlier cluster to be selected.
This small experiment suggests that we can potentially extend our approach to handle multi-prototype shape families by returning the largest clusters that satisfy certain statistical relationships with one another.
Future work should more comprehensively evaluate this on a larger set of multi-prototype prompts and explore other decision criteria for selecting multiple clusters.}

% well-define shape with multiple modes.
%
% there are shapes like letters (either much more prototypes or more sophisticated warps).
%
% in this work, we should its enough to have 1 proto + common warps to cover a large set of motion trajectories.
%
% \maneesh{Ideally a figure showing this limitation of the heptagram should be included as most readers probably won't know what the two forms really area. Ideally in supplemental materials we should show some examples of prompts that have more than 1 prototype.}

% \jiaju{
% - show an example of Ws? 
% - another experiment with heptagrams and Ws
% }

\begin{figure}[tb]
    \centering
    \includegraphics[width=\linewidth]{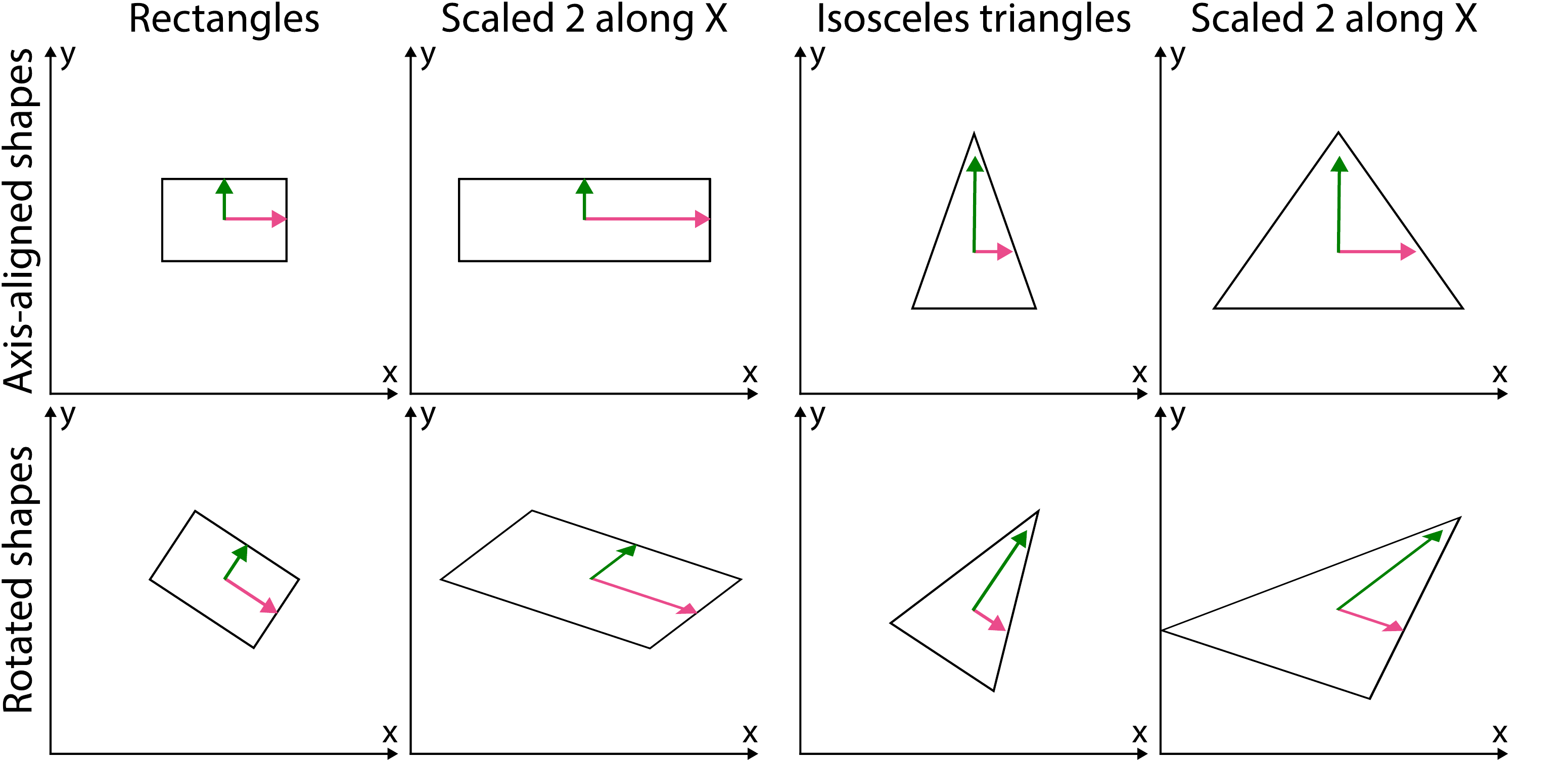}
    \caption{When a shape is aligned with the screen space's coordinate frame, applying an anisotropic scale of 2 along the x-axis does not introduce skews to the shapes (top). However, once a shape is rotated out of alignment, non-uniform scales effectively skew the shapes out of their respective shape families (bottom).
    % \kenny{Make a bigger space between them}
    }
    \label{fig:anisotropic}
\end{figure}

\subsection{Shape Families with Anisotropic Similarity}
% \maneesh{This paragraph needs to be its own subsection. Needs to also come earlier before failure modes. It is not about failure modes. This is describing an important aspect of our shape families.}
% \jiaju{shape-agnostic}
% As demonstrated in Figure~\ref{fig:anisotropic}, f
For shape families with the anisotropic similarity warp,
% (e.g., rectangles, ellipses, isosceles triangles, etc), 
when a shape is not aligned with the screen coordinate frame (bottom row of Figure~\ref{fig:anisotropic}),
a skew can be introduced to distort it out of its family if a non-uniform scaling is applied to it~\cite{steger2012least}.
For symmetric shapes like the ones shown in Figure~\ref{fig:anisotropic}, their lines of symmetry need to align with the screen coordinate axes.
For asymmetric shapes with right angles (e.g., right triangles), the lines along the right angles need to be in alignment so that they are preserved under scaling.
This means that, in order to properly warp to all other members of the family, we need a prototype that is in alignment with the screen space's coordinate frame for non-uniform scaling to be applied without skew.
We manually ensure this when we are constructing ground truth shape families for our dataset, and assume that the LLM will generate some trajectories in such canonical orientations.

% \maneesh{I think a discussion of isosceles triangles and right angles and right triangles would probably be good to add in.}

% The definition of principal axes varies from shape to shape, so the initial orientation of the prototype trajectory being warped is important.

%% file: suppl_sec/failure_modes.tex
\section{Failure Modes}
\label{sec:supp_lim}

\input{suppl_sec/table_decision_criteria}

\begin{figure*}[t]
    \centering
    \includegraphics[width=\textwidth]{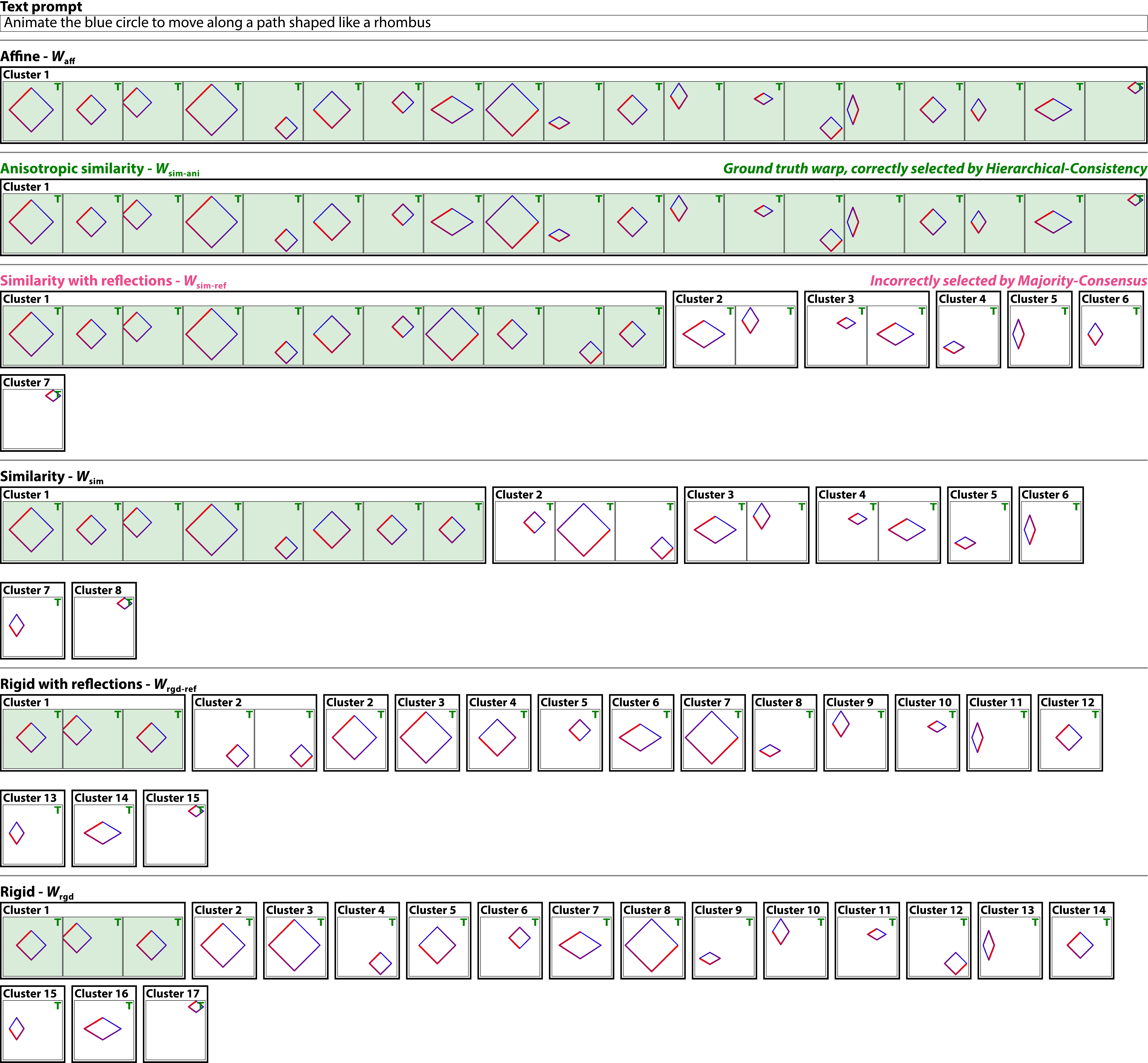}
    \caption{We demonstrate a case where \DChier{} chooses the correct warp but \DCmajor{} fails by selecting a more restrictive warp.
    The prompt (top) asks for a rhombus-shaped path.
    We show clustering results of all warps in our hierarchy, with largest clusters under each warp highlighted in green. 
    Each trajectory has a ground truth correctness label on the upper right.}
    \label{fig:hierarchy_rhombus}
    \vspace{-3mm}
\end{figure*}

\begin{figure*}[t]
    \centering
    \includegraphics[width=\textwidth]{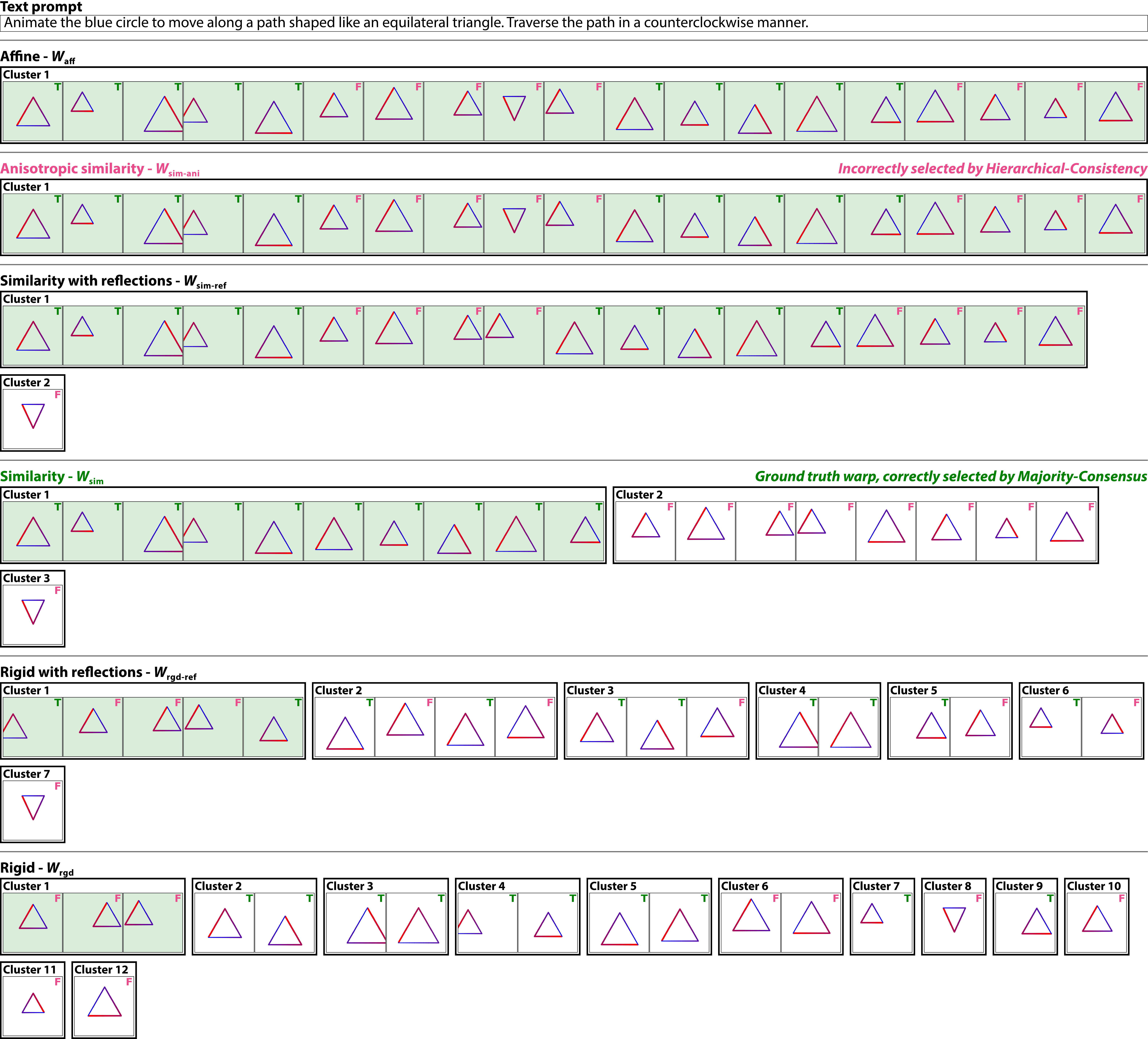}
    \caption{An example case where \DCmajor{} correctly chooses the ground truth warp but \DChier{} fails by selecting a less restrictive warp.
    The prompt (top) asks for a path shaped like an equilateral triangle traversed in a counterclockwise manner.
    We show clustering results of all warps in our hierarchy.
    % \maneesh{Say what the shape of the trajectory is supposed to be here in the caption for this Fig and all all Fig's showing trajectories.} %
    }
    \label{fig:hierarchy_triangle}
    \vspace{-3mm}
\end{figure*}

\subsection{Decision Criteria Failure Modes}
\label{sec:supp_dc_failure}

% \maneesh{I'm confused by this introduction. I don't think we need to reiterate about the graound truth W. Also The decision criteria are based on assumptions. In the next sentence, what does it mean to say "When these assumptions are not made ..." Who is making the assumptions, or not making them?}
% \kenny{Addressed}

Our method can recover a shape family from an input prompt by using some decision criteria to estimate an appropriate transformation group $W$.
In Section 4.3 of the main paper, we introduce two such decision criteria that analyze the distribution of motion trajectories generated by an LLM.
When different sets of assumptions are met, in terms of how the LLM generates motion trajectories, these decision criteria will predict the correct transformation group.
For \DCmajor~the assumptions it makes are (i) the LLM produces correct trajectories more frequently than incorrect ones; (ii) the collection of sampled trajectories diversely covers the modes of variation possible within the transformation group.
\DChier~relaxes this second assumption, but introduces a new assumption in that incorrect trajectories produced by the LLM will not cluster with the correct trajectories produced by the LLM under any transformation group in our hierarchy.
In cases where these assumptions are broken, it is possible for these decision criteria to fail (i.e. recover a different transformation group compared with the ground-truth annotation).

To demonstrate such cases, we include qualitative examples in Figure~\ref{fig:hierarchy_rhombus} and Figure~\ref{fig:hierarchy_triangle}.
In each figure, we show LLM sampled motion trajectories from a text prompt (top of figure), and how these motion trajectories are clustered under different transformation groups from our hierarchy. 
The top transformation group we show is affine, the least restrictive, and transformation groups get progressively more restrictive, ending with rigid.
For each transformation group, we highlight the largest cluster in green.
Each trajectory is labeled with a T/F value in the top-right corner, indicating whether it is a member of our ground-truth shape family.

In Figure~\ref{fig:hierarchy_rhombus}, we demonstrate a case where \DChier{} identifies the correct $W$, while \DCmajor{} fails.
For this prompt, all of the LLM generated motion trajectories were correct.
For \DChier, for the affine transformation group it identifies the largest cluster, and observes that this cluster does not lose any members when the transformation group is made more restrictive with anisotropic similarity, which is the correct transformation group for this shape family. 
Trying to make $W$ more restrictive (similarity with reflections) would break this cluster, so \DChier{} succeeds by stopping at the previous step.
In contrast, \DCmajor{} identifies that under the clustering produced by similarity with reflections, over half of the generations would be included in the largest cluster, so it incorrectly would choose an overly restrictive transformation group.
This sampling breaks an assumption made by~\DCmajor, as the LLM produced trajectories do not provide a balanced enough coverage over the modes of variation under the ground-truth transformation group (anisotropic similarity).

In Figure~\ref{fig:hierarchy_triangle}, we demonstrate a case where \DCmajor{} identifies the correct $W$, while \DChier{} fails.
In this case, 10 LLM generations were correct and 9 LLM generations were incorrect, as they produced paths that did not traverse in a clockwise manner (note time is indicated as a progression from blue to red in the renders).
The correct transformation group in this case is similarity, as more permissive transformation groups would cause matches that violate the traversal orientation specification.
\DCmajor{} starts with the clustering produced by the most restrictive transformation group (rigid), which has a largest cluster of size 3, and progressively considers less restrictive transformation groups until it finds a cluster with at least 10 members, correctly identifying similarity should be used as $W$.
In contrast,~\DChier{} starts with affine, and once again identifies that all of the LLM productions are grouped into a single cluster.
This cluster only loses members on the transition from anisotropic similarity to similarity with reflections, so \DChier{} will incorrectly choose an overly permissive transformation group.
In this case, some of the incorrect LLM samples would cluster with correct samples under the affine transformation group, breaking an assumption made by \DChier.

% \maneesh{Next paragraph has some grammatical errors in it.}
% \maneesh{What is the failure mode. I'm confused here.}
% \kenny{Addressed these}

So, depending on the distribution of the LLM produced motion trajectories, these decision criteria can fail in their task of recovering the correct transformation group.
We quantitatively analyze the prevalence of these failure modes in Table~\ref{tab:heir_breakdown}. 
For our two proposed criteria, \DChier~and \DCmajor, we record for all 224 prompts how the chosen transformation group compares with the ground-truth transformation group in the benchmark dataset.
\textit{Match} indicates the criteria chose the correct transformation group, while \textit{more restrictive} indicates the criteria chose a transformation group too low in the hierarchy, and \textit{less restrictive} indicates the criteria chose a transformation group too high in the hierarchy. 
% \maneesh{Need to show examples of both cases in figures and describe what happens -- why the DC are wrong -- before analyzing table results. More restrictive means that the samples did not exercise the full family or some such. Similarly explain what less restrictive means -- both in terms of concrete example in figure.}
% \kenny{I think this is covered now}

In general, we find that \DChier{} more reliably identifies the correct $W$ compared with \DCmajor{} (72\% vs 49\%).  
% \maneesh{Need to add a sentence or two explaining why before going on to the next parts.}
% \kenny{Addressed}
This likely indicates that the assumptions made by \DChier~are more often met compared with the assumptions made by \DCmajor.
From the qualitative examples, we can see the LLM struggles to produce sufficiently diverse samples that cover all of the possible modes of variation, so this is likely a major source of error for \DCmajor.
Beyond checking whether or not these decision criteria recover the exact ground-truth $W$, we can additionally analyze how these methods tend to fail.
\DCmajor~starts at the bottom of the hierarchy, stopping when a majority clustering is found.
As such, \DCmajor~is overly prone to producing more restrictive transformation groups when it makes mistakes (25 times more likely to have a more restrictive error compared with a less restrictive error).
\DChier~starts at the top of the hierarchy, with a stopping criterion when the majority cluster begins to break apart.
As such, \DChier~is overly prone to producing transformation groups that are too loose when it makes mistakes (5 times more likely to have a less restrictive error compared with a more restrictive error).
These quantitative trends match the qualitative examples.

% \jiaju{[high] how clusters evolve under different warps for one or two}
% ideally one that both decision criteria fails on.
% simple majority failL - 1107_m_rb_rhombus
% hierarchical fail - 1107_m_pol_tr_dir_equilateral_triangle_tracing_direction_ccw

\begin{figure*}[t]
    \centering
    \includegraphics[width=\textwidth]{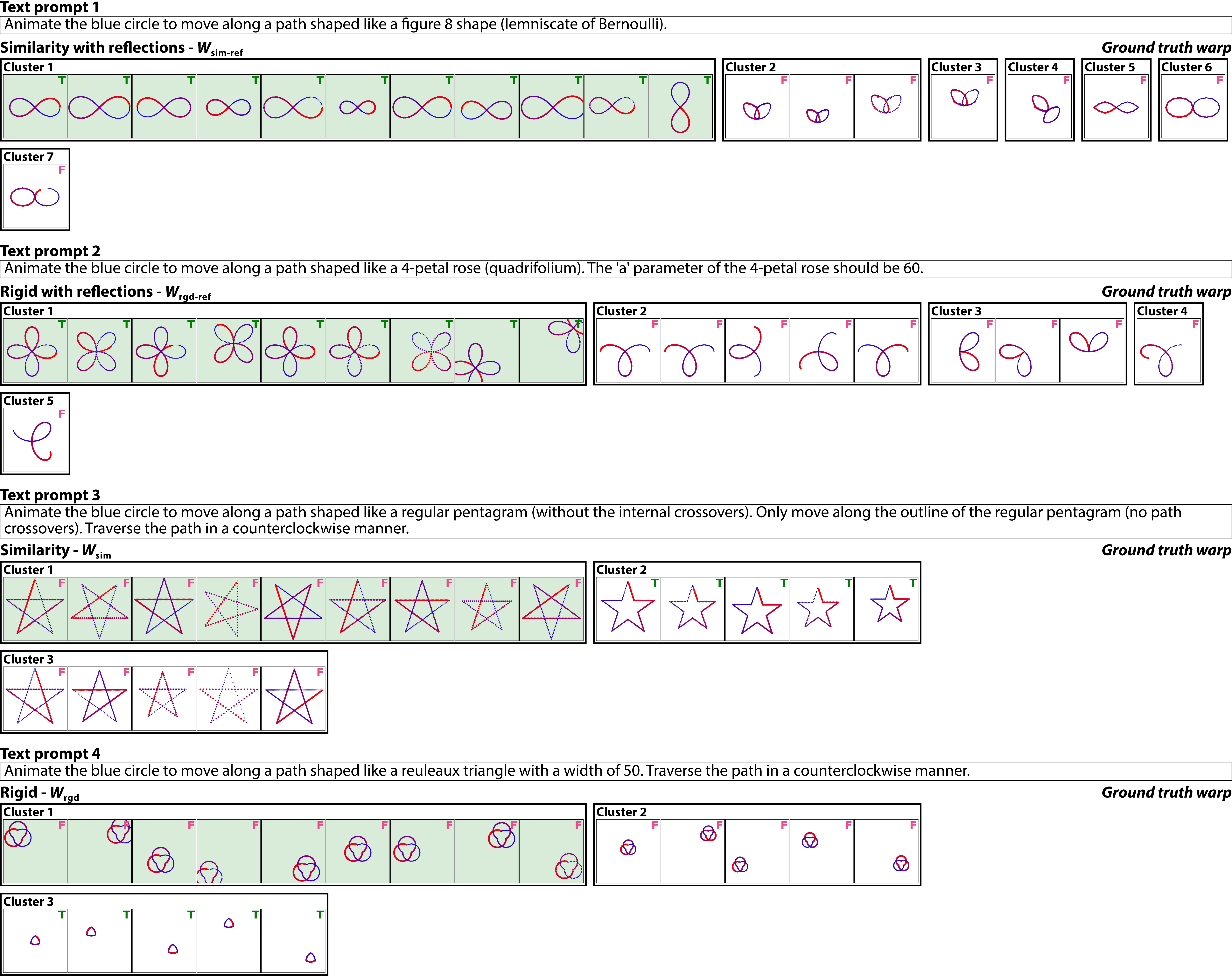}
    \caption{More results of applying our self-consistency approach with the oracle warps. The top half shows two success cases where the largest clusters do correspond to the correct shape families (figure 8 and 4-petal rose).
    The bottom half shows two failure cases where the LLM did not produce more members of the ground truth shape families than non-members (pentagram outline and reuleaux triangle).
    }
    \label{fig:results_true_false}
    \vspace{-3mm}
\end{figure*}

\subsection{LLM Failure Modes}
% [high] topic: analysis of failure modes of LLM generations, under ground-truth warp
%
The general self-consistency approach makes an assumption that the LLM is likely to produce the correct response more often than any other incorrect response~\cite{wang2022self}.
We examine cases where this assumption does not hold and present quantitative results in Figure~\ref{fig:results_true_false}.
Applying our self-consistency approach with the oracle warps to the trajectories generated by GPT-5 from the 224 prompts in our dataset (same experimental set-up as Section 5.2 of the main paper), we see 16.1\% of the prompts with largest clusters that do not contain any correct trajectories.
For GPT-4.1 generated trajectories, we have 31.3\% (70) such clusters.
% \maneesh{Need to explain what these numbers mean. I don't follow what the prev sentence is trying to say.}
%
% on GPT-5 diversity samples: 188 / 224 (83.93\%) clusters chosen has correct trajectories
% on GPT-4.1 diversity samples: 154 / 224 (68.75\%) clusters chosen has correct trajectories

We show examples of these failure cases in the bottom half of Figure~\ref{fig:results_true_false}.
Text prompt 3 asks for a trajectory in the shape of an outline of a pentagram, and the oracle similarity warp puts the generated trajectories into three clusters.
While the LLM (GPT-5) was able to produce the correct trajectory 5 times (Cluster 2), it generates more instances of a regular pentagram shape with the internal path crossovers in both clockwise and counterclockwise fashion.
Text prompt 4 asks for counterclockwise reuleaux triangle-shaped trajectories.
Under the ground truth rigid transformation warp, the generated trajectories are grouped into three clusters.
The LLM (GPT-4.1) similarly generates the correct trajectory 5 times (Cluster 3), but falsely traces the construction lines of a reuleaux triangle for the other 14 times.
In both of these cases, the incorrect trajectories formed largest clusters in the form of relative majority with 9 trajectories.
As the LLM produces more trajectories outside of the ground truth shape families than inside, the general assumption of self-consistency does not hold and the largest clusters do not contain any correct trajectories.

%% file: suppl_sec/table_decision_criteria.tex
\begin{table}[t]
\centering
\caption{Our decision criteria has different failure modes when not selecting the ground truth warp. 
\DCmajor{} selects more restrictive distance metrics for 108 prompts (95.6\% of the time when it is not matching with the ground truth), while \DChier{} is more conservative for 50 prompts (80.6\% of the time). 
% \DCmajor{} selects more restrictive distance metrics for 95.6\% of the time, while \DChier{} is more conservative for 80.6\%. 
Note that both decision criteria choose distance metrics that are on the same level of the hierarchy but different from the ground truth for 1 and 3 prompts respectively.
% \kenny{Maybe we just show percentages in table?}
% \kenny{Change orientation -- mention equal in caption}
}
\resizebox{\columnwidth}{!}{%
\begin{tabular}{@{}lrrr@{}}
\toprule
Decision Criteria & More Restrictive & Match & Less Restrictive \\
\midrule
\DCmajor{} & 48.2\% & 49.6\% & 1.8\% \\
\DChier{}  & 4.0\%  & 72.3\% & 22.3\% \\
\bottomrule
\end{tabular}%
}
\label{tab:heir_breakdown}
\end{table}

% \begin{tabular}{@{}lrrr@{}}
% \toprule
% Decision Criteria & More Restrictive & Match & Less Restrictive \\
% \midrule
% \DCmajor{} & 108 (48.2\%) & 111 (49.6\%) & 4 (1.8\%) \\
% \DChier{} & 9 (4.0\%) & 162 (72.3\%) & 50 (22.3\%) \\
% \bottomrule
% \end{tabular}%

% \begin{table}[t]
% \centering
% \caption{Our decision criteria has different failure modes when not selecting the ground truth warp. \DCmajor{} selects more restrictive distance metrics for 95.6\% of the time, while \DChier{} is more conservative for 80.6\%.\kenny{Change orientation -- mention equal in caption}}
% \resizebox{\columnwidth}{!}{%
% \begin{tabular}{@{}lrr@{}}
% \toprule
% Category & \DCmajor{} & \DChier{} \\
% \midrule
% Match & 111 (49.6\%) & 162 (72.3\%) \\
% Non-match & 113 (50.4\%) & 62 (27.7\%) \\
% \cmidrule(lr){1-3}
% \quad More Restrictive & \textbf{108 (95.6\%)} & 9 (14.5\%) \\
% \quad More Permissive & 4 (3.5\%) & \textbf{50 (80.6\%)} \\
% \quad Equal & 1 (0.9\%) & 3 (4.8\%) \\
% \bottomrule
% \end{tabular}%
% }
% \label{tab:heir_breakdown}
% \end{table}

%% file: suppl_sec/method_details.tex
%New colors defined below
\definecolor{codegreen}{rgb}{0,0.6,0}
\definecolor{codegray}{rgb}{0.5,0.5,0.5}
\definecolor{codepurple}{rgb}{0.58,0,0.82}
\definecolor{backcolour}{rgb}{0.95,0.95,0.92}

%Code listing style named "mystyle"
\lstdefinestyle{mystyle}{
  backgroundcolor=\color{backcolour},   
  commentstyle=\color{codegreen},
  keywordstyle=\color{magenta},
  numberstyle=\tiny\color{codegray},
  stringstyle=\color{codepurple},
  basicstyle=\ttfamily\footnotesize,
  breakatwhitespace=false,         
  breaklines=true,                 
  captionpos=b,                    
  keepspaces=true,                 
  numbers=left,                    
  numbersep=5pt,                  
  showspaces=false,                
  showstringspaces=false,
  showtabs=false,                  
  tabsize=2,
  frame=ltb,
  framerule=0pt,
  % lineskip=-10pt, % Adjust this value if needed
}

\lstset{style=mystyle}

\section{Method Implementation Details}
\label{sec:supp_method}

\subsection{Computing Warp-Invariant Distance}
\input{suppl_sec/pseudocode_icp}
% We provide implementation details for the warp-invariant distance metrics
% described in Section 3 of the main paper.
Algorithm~\ref{alg:icp} presents our ICP-based approach~\cite{besl1992method,chen1992object,du2010affine} for computing the warp-invariant distance between a source trajectory $R$ and target trajectory $T$ under a geometric warp $W$.
%
% The algorithm consists of two nested loops.
% The outer loop samples random subsets of point correspondences to robustly handle potential misalignments.
% The inner loop performs iterative refinement to find the optimal transformation $A$ that minimizes the point-to-point distance between the transformed source and target trajectories.
%
% \paragraph{Outer loop: Random sampling.}
We begin by resampling both trajectories to have $n$ equally-spaced points along their arc length. This resampling establishes initial point correspondences, where we assume that points at the same index in $R$ and $T$ correspond to one another. The outer loop then randomly selects 3 non-repeated, non-collinear point indices $\mathcal{I}$ from these trajectories. These three points provide sufficient constraints to estimate most geometric transformations in our hierarchy.

% \paragraph{Inner loop: ICP refinement.} 
Given the sampled point indices, the inner loop iteratively refines the transformation $A$. At each iteration, we: (1) estimate a transformation $A$ that aligns the sampled points from $R_{\text{trans}}$ to the corresponding points in $T$, (2) apply this transformation to the source trajectory $R$, (3) resample the transformed trajectory for a new set of corresponding points, and (4) compute the average point-to-point distance. If the change in distance falls below a convergence threshold $\epsilon$, the inner loop terminates. Otherwise, we map the newly sampled points back to the original $R$ space by undoing the transformation and repeat.
This iterative process effectively ``nudges'' points along trajectory $R$ to better align with $T$ under the transformation $A$.

After convergence, we check if the current alignment achieves a lower distance than previous iterations and update the best transformation $A_{\min}$ accordingly. The algorithm terminates when the distance falls below a threshold $\tau$ or after a maximum number of outer iterations.
\camready{Our ICP-based implementation is a proof of concept to show the viability of our self-consistency approach for motion trajectories. It would likely be possible to improve its runtime with efficient variants in the future~\cite{rusinkiewicz2001efficient,mellado2014super}.}

\paragraph{Transformation estimation.} 
\texttt{EstimateTransform} in line 11 of Algorithm~\ref{alg:icp} computes $A$ based on the geometric warp type $\mathcal{W}$.
For rigid and similarity transformations (with or without reflections), we use the Kabsch-Umeyama algorithm~\cite{kabsch1976solution,umeyama1991least}, which computes optimal rotation, translation, and optionally uniform scaling and reflection via SVD.
For affine warps, we directly solve the linear system relating source to target points.
% —three correspondences suffice to determine the six affine parameters. 
For anisotropic similarity, we test both $A = M_{\text{trs}} \cdot M_{\text{rot}} \cdot M_{\text{scl}}$ and $A = M_{\text{trs}} \cdot M_{\text{scl}} \cdot M_{\text{rot}}$ decomposition orders and select the transformation yielding the smaller distance.

\paragraph{Handling closed trajectories}
Our assumption that points with the same indices after arc-length resampling correspond to one another holds for open trajectories with distinct starting and end points.
However, closed trajectories (e.g., circles, squares) may have arbitrary starting points.
This means that, if $T$ and $R$ are the same closed trajectories but with distinct starting points, our algorithm will not be able to find an $A$ that produces zero distance between them.
% A square path that starts at a corner produces a different parameterization than one that starts along a side, even though they represent the same geometric shape.
%
To address this, we extend our algorithm with a coarse-to-fine search over different starting points along $R$. We compute the distance metric with each point in $R$ taken as the starting point and report the minimum distance across all starting points.

\begin{table}[t]
\centering
\caption{F1 scores of verification by \DChier{} and the average number of DBSCAN clusters across the distance threshold $\tau$.}
\vspace{-3mm}
\label{tab:distance_threshold}
\resizebox{\columnwidth}{!}{%
\begin{tabular}{lccccccc}
\toprule
& $\tau=0.25$ & $\tau=0.5$ & $\tau=1.0$ & $\tau=2.0$ & $\tau=4.0$ & $\tau=8.0$ \\
\midrule
F1 score & 84.0 & 84.6 & 83.8 & 83.1 & 80.3 & 76.8 \\
\midrule
Clusters (avg$\pm$std) & 4.0$\pm$4.9 & 3.7$\pm$4.8 & 3.5$\pm$4.5 & 3.1$\pm$4.0 & 2.6$\pm$3.1 & 2.1$\pm$2.1 \\
\bottomrule
\end{tabular}%
}
\end{table}

\subsection{Sensitivity to DBSCAN Distance Threshold}
\label{sec:supp_dbscan}

\camready{Equation 2 of the main paper defines that two trajectories in the same shape family should have a distance of 0.
We perform DBSCAN clustering with a small threshold $\tau$ (Section 4.2 of the main paper) only to account for discretization errors during resampling in our distance metric computation.
Our approach is not sensitive to the value of $\tau$. As shown in Table~\ref{tab:distance_threshold}, increasing $\tau$ by a factor of 32 (from 0.25 to 8.0) results in only a 7.2 percentage point drop in F1 score.}

\subsection{Generating Motion Trajectories with an LLM}
\label{sec:supp_gen_llm}
% - [high] more details on our prompting strategy for motion synthesis programs + diversity (our prompts)~\cite{ma2025mover}

To use an LLM for motion trajectory animation generation, we follow the approach proposed by MoVer~\cite{ma2025mover};
% \footnote{MoVer repository: \url{https://github.com/jama1017/MoVer}};
we prompt an LLM with a system message containing overall instructions and the complete documentation of an animation API (GSAP~\cite{gsap}), followed by a user prompt that contains the static SVG code to be animated, the description of a motion trajectory, and additional instructions.

We modify the MoVer approach in two places.
First, we updated the animation API with an additional function \texttt{setProperty()} that allows changing an element's properties like positions without adding animation tweens.
\begin{lstlisting}
/**
 * This function immediately sets properties on an SVG element without animation. Use this function to initialize or instantly update element properties such as position, scale, rotation, opacity, etc. This is useful for setting up initial states before animating.
 * @param {object} element - The SVG element whose properties should be set.
 * @param {object} properties - An object containing property-value pairs to set on the element. Common properties include x, y, scaleX, scaleY, rotation, opacity, transformOrigin, and svgOrigin.
 * @returns {void} - This function does not return anything.
 * @example
 * // Set the translate transform value of the element to [100, 200].
 * // Here, x, y values set the translation of the element (displacement from its original position at the start of the animation).
 * setProperty(circle, { x: 100, y: 200 });
 */
function setProperty(element, properties)
\end{lstlisting}

For the user prompt, we append additional instructions shown below after the motion trajectory prompt. At the end, we instruct the LLM to produce five diverse animations following the diversity prompting strategy~\cite{zhang2025verbalized}.
\begin{lstlisting}
SVG Code:
{{ svg code }}

Prompt:
{{ motion trajectory prompt }}

## Additional Instructions
- Note that the negative y-axis is the upward direction in the SVG scene.
- To start a path animation, use setProperty() to position the element at the beginning of the path. Be mindful of the SVG viewbox and not move the element and the path out of view.
- For curved paths, use multiple consecutive very short linear translations to achieve smooth motions.
- The entire animation should last between 1 and 5 seconds unless otherwise specified.

Generate 5 animations to the prompt, each within a separate ```javascript ``` tag.
Each animation must include a numeric probability value as a comment within its code block.
Please sample at random from the tails of the distribution, such that the probability of each response is less than 0.10.
\end{lstlisting}

\begin{figure}[t]
    \centering
    \includegraphics[width=0.70\linewidth]{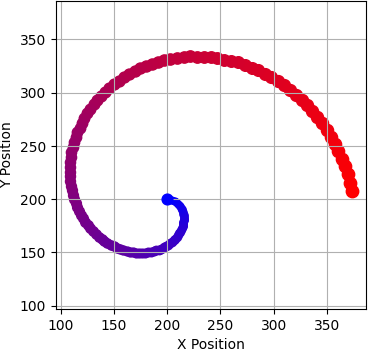}
    \caption{We prompt a VLM with a scatter plot representation of a motion trajectory. Each dot is the center position of the animated element and the color indicates time (transition from blue to red).
    % Animate the blue circle to move along a path shaped like an archimedean spiral (initial radius should be 0 and move outward from the scene center). Complete 1 turn of the archimedean spiral.%
    }
    \label{fig:vlm_plot}
    \vspace{-5mm}
\end{figure}

\subsection{Verifying Motion Trajectories with a VLM}
\label{sec:supp_verify_vlm}
% - [high] more details on the VLM baseline verifier (prompt + figure of what we send to VLM for verification)

Our experiments use two VLMs (GPT-4.1 and GPT-5) as baseline verification methods, and we provide details here on our prompting approach.
We represent the motion trajectory center point time series data of animation as a scatter plot where the color of each dot transitions from blue to red to indicate the passage of time (Figure~\ref{fig:vlm_plot}).

We prompt a VLM with a system message asking it to output a true/false label indicating whether a trajectory animation follows a trajectory prompt or not, with instructions on how to read the scatter plot.
We then add a user message containing the motion trajectory prompt and the plot.
\begin{lstlisting}
You are an expert in evaluating how well motion trajectories follow given prompts. Given a plot of a motion trajectory, your task is to assess whether the trajectory follows the prompt (True or False). Blue indicates the beginning of the trajectory and red indicates the end of the trajectory. The color transitions from blue to red as the trajectory progresses across time.

Please output your response as a JSON object with the following format:
```json
{
    "<trajectory_id (the index of the trajectory file)>": 0 or 1, # 0 means False, 1 means True
}
```
\end{lstlisting}
% \maneesh{People will likely miss figures coming after references. I would integrate them into the main part of this document rather than putting them after the refs.}
% User prompt: the animation prompt + trajectory figure
% \jiaju{TODO: add trajectory plot as a figure}

%% file: suppl_sec/pseudocode_icp.tex
\begin{algorithm}[t]
\caption{ICP-based Trajectory Distance Metric}
\label{alg:icp}
\begin{algorithmic}[1]
\State \textbf{Input}: Source trajectory $R$, target trajectory $T$, and geometric warp type $\mathcal{W}$
\State \textbf{Output}: Minimum distance $d_{\min}$ between $R$ and $T$ with corresponding transformation matrix $A_{\min}$\\

\State $T \gets \text{ResampleByArcLength}(T, n)$
\State $d_{\min} \gets \infty$, $A_{\min} \gets \mathbf{I}$

\For{$i = 1$ to $M_{\text{outer}}$}
    \State $R_{\text{trans}} \gets \text{ResampleByArcLength}(R, n)$
    \State $d_{\text{prev}} \gets \infty$
    \State $\mathcal{I} \gets \text{Randomly sample 3 point indices}$
    
    \For{$j = 1$ to $M_{\text{inner}}$}
        \State $A \gets \text{EstimateTransform}(R_{\text{trans}}[\mathcal{I}], T[\mathcal{I}], \mathcal{W})$
        \State $R_{\text{trans}} \gets A \cdot R$ \Comment{Apply transformation}
        \State $R_{\text{trans}} \gets \text{ResampleByArcLength}(R_{\text{trans}}, n)$
        
        \State $d_{\text{avg}} \gets \frac{1}{n}\sum_{k=1}^{n} \|R_{\text{trans}}[k] - T[k]\|_2$
        
        \If{$|d_{\text{avg}} - d_{\text{prev}}| < \epsilon$}
            \State \textbf{break} \Comment{Inner loop converged}
        \EndIf
        
        \State $d_{\text{prev}} \gets d_{\text{avg}}$
        \State $R_{\text{trans}} \gets A^{-1} \cdot R_{\text{trans}}$ \Comment{Undo transformation}
    \EndFor
    
    \If{$d_{\text{avg}} < d_{\min}$}
        \State $A_{\min} \gets A$, $d_{\min} \gets d_{\text{avg}}$ \Comment{Update alignment}
    \EndIf
    
    \If{$d_{\text{avg}} < \tau$}
        \State \textbf{break} \Comment{Outer loop converged}
    \EndIf
\EndFor
\State \Return $d_{\min}$, $A_{\min}$
\end{algorithmic}
\end{algorithm}

% max outer iterations $M_{\text{outer}}$, max inner iterations $M_{\text{inner}}$, convergence threshold $\epsilon$, RANSAC threshold $\tau$, n
% non-repeated, non-collinear

%% file: suppl_sec/benchmark_figures.tex
\clearpage
\begin{figure*}[t]
    \centering
    \includegraphics[width=\textwidth]{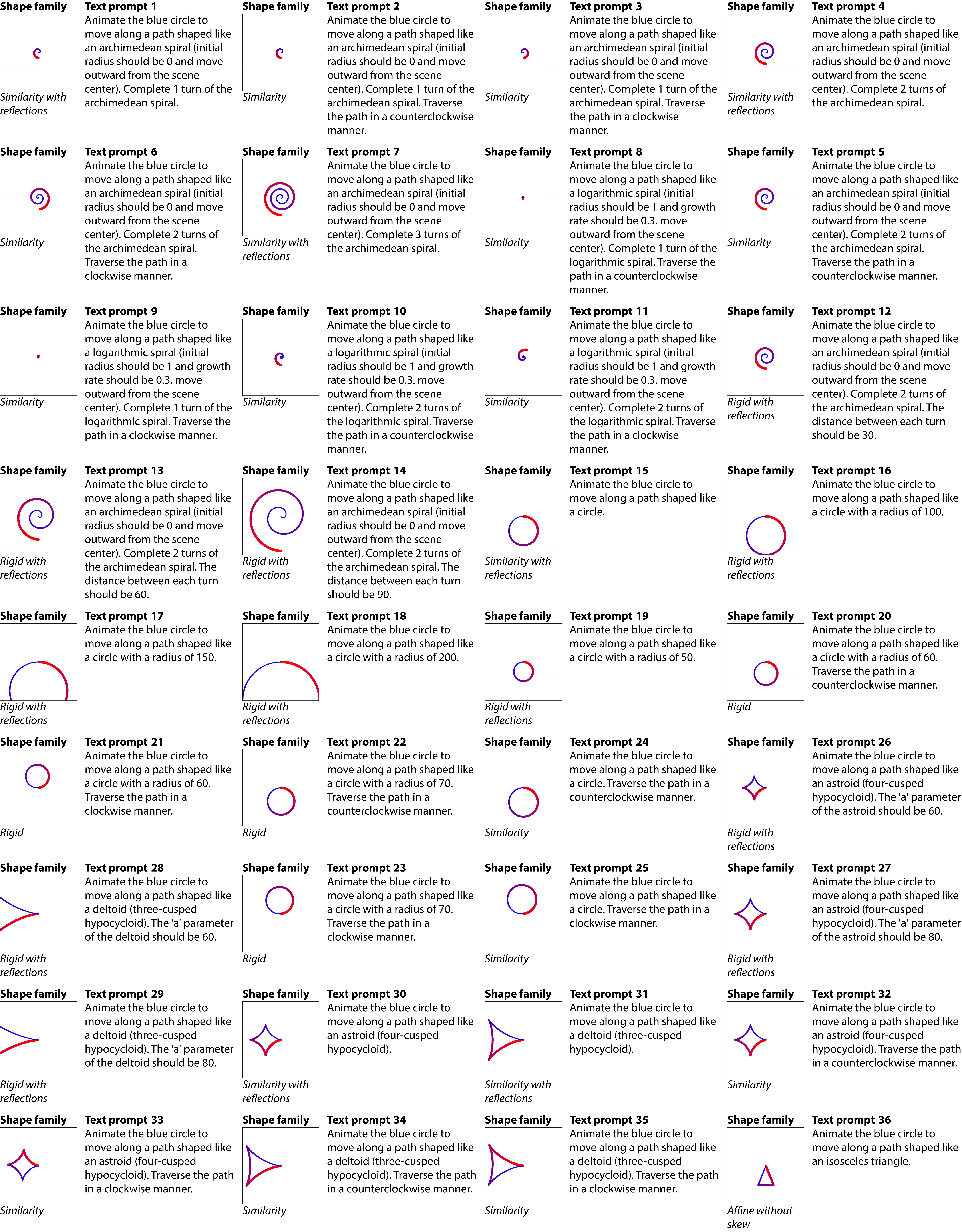}
    \caption{Motion trajectory benchmark: prompt 1--36 with ground truth shape families.}
    \label{fig:bench_mark_1}
\end{figure*}

% \section{All Prompts and Shape Families}
% Here we present all 224 prompts in our benchmark with ground truth shape families.

\begin{figure*}[t]
    \centering
    \includegraphics[width=\textwidth]{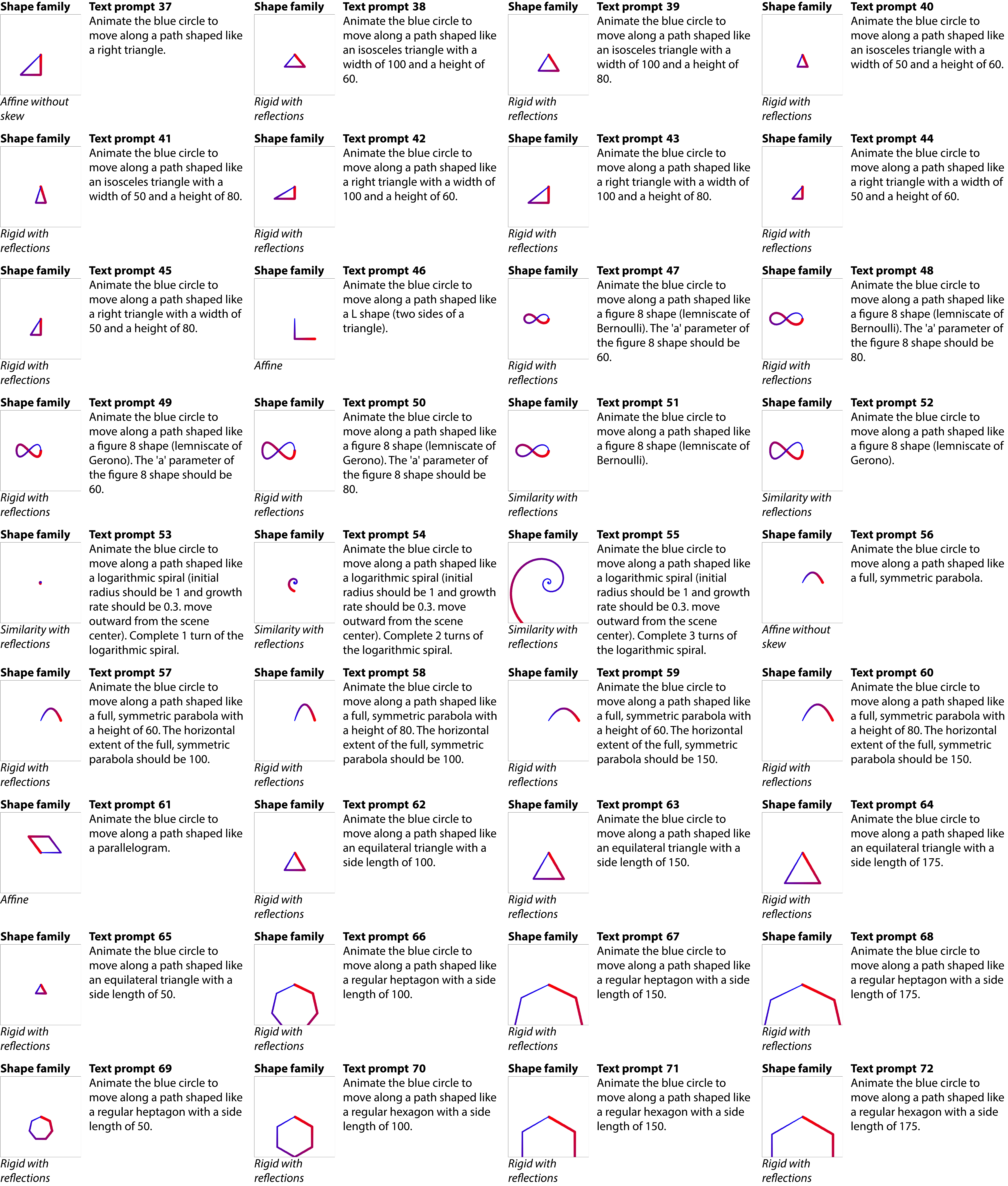}
    \caption{Motion trajectory benchmark: prompt 37--72 with ground truth shape families.}
    \label{fig:bench_mark_2}
\end{figure*}

\begin{figure*}[t]
    \centering
    \includegraphics[width=\textwidth]{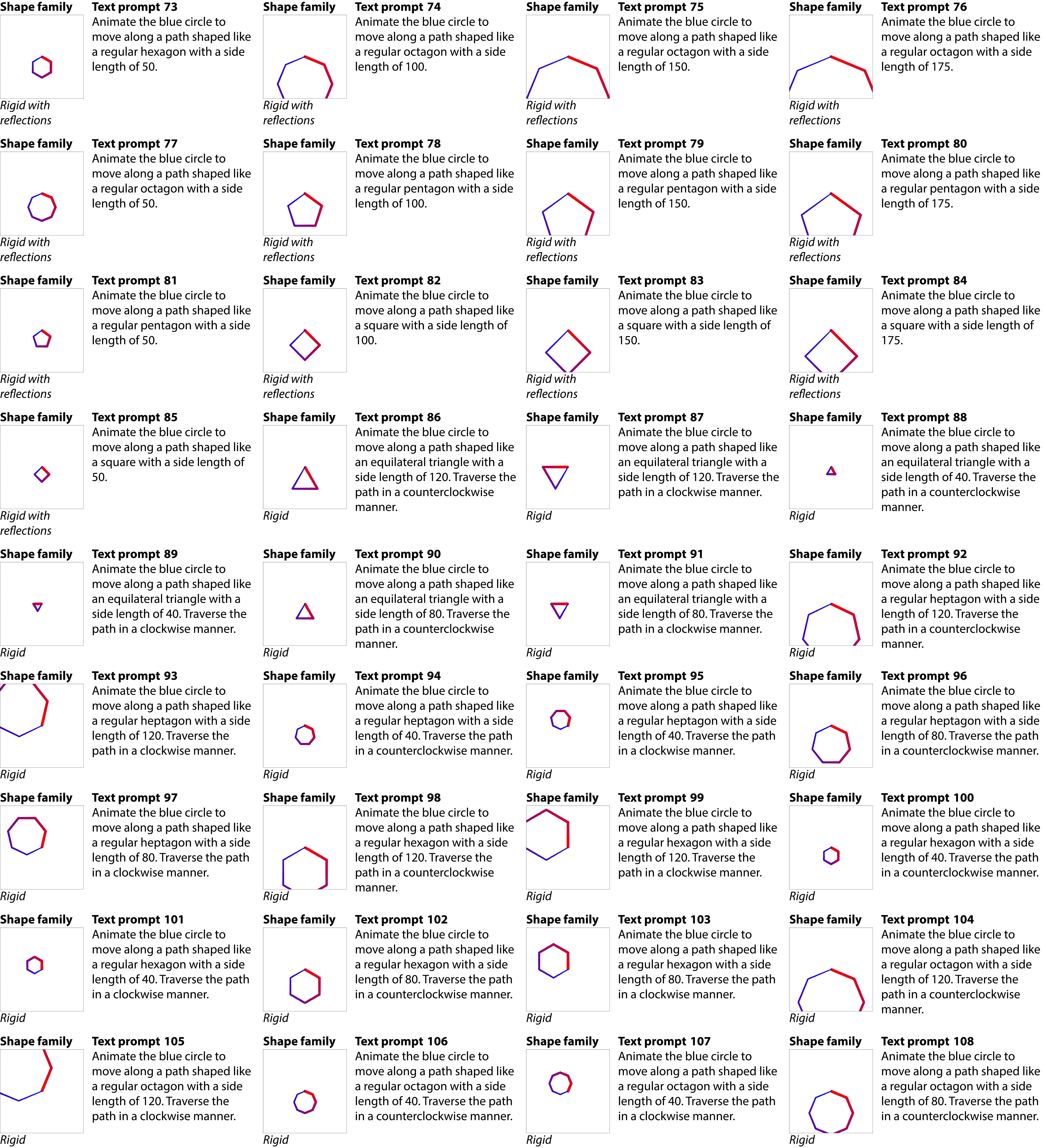}
    \caption{Motion trajectory benchmark: prompt 73--108 with ground truth shape families.}
    \label{fig:bench_mark_3}
\end{figure*}

\begin{figure*}[t]
    \centering
    \includegraphics[width=\textwidth]{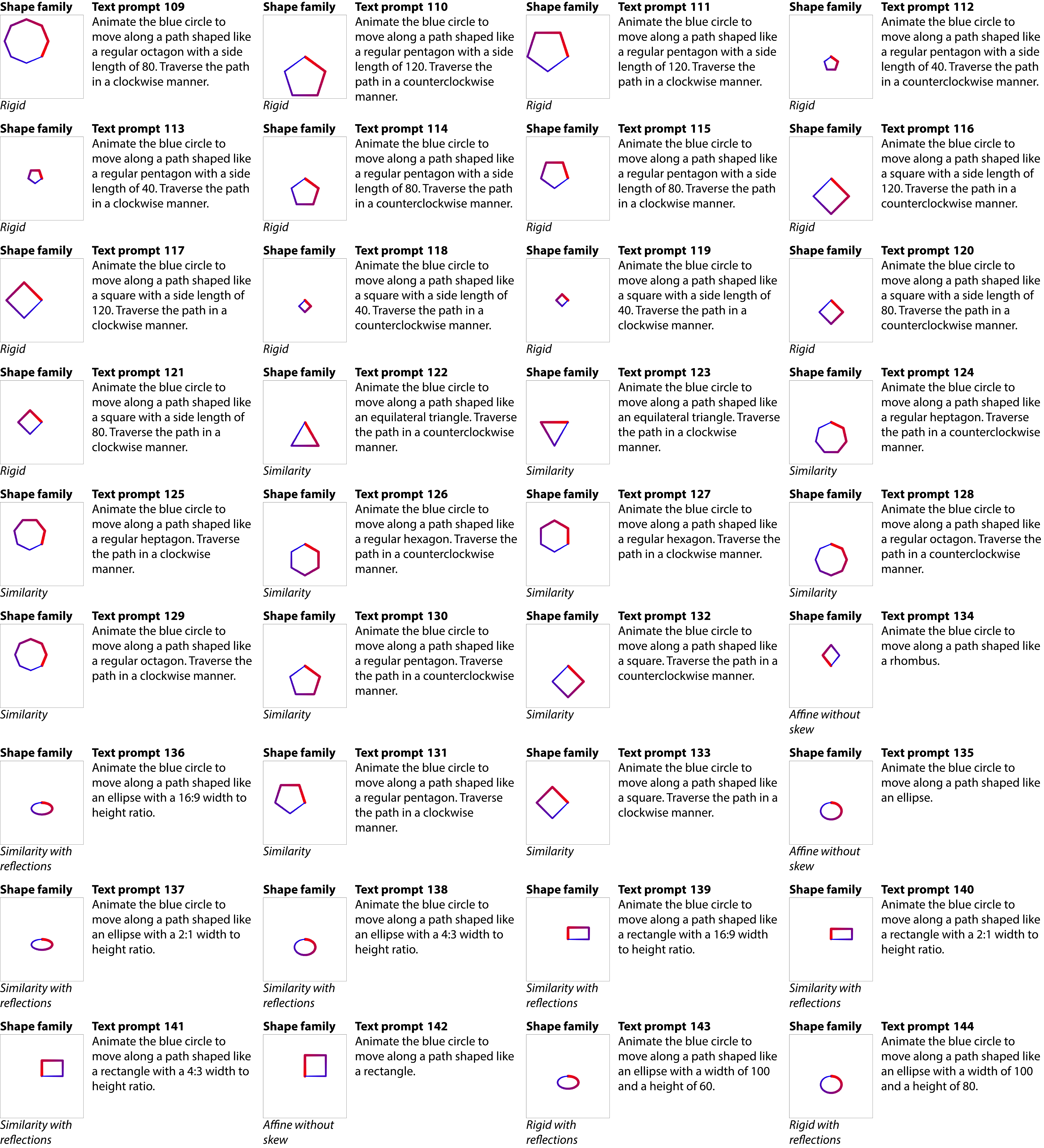}
    \caption{Motion trajectory benchmark: prompt 109--144 with ground truth shape families.}
    \label{fig:bench_mark_4}
\end{figure*}

\begin{figure*}[t]
    \centering
    \includegraphics[width=\textwidth]{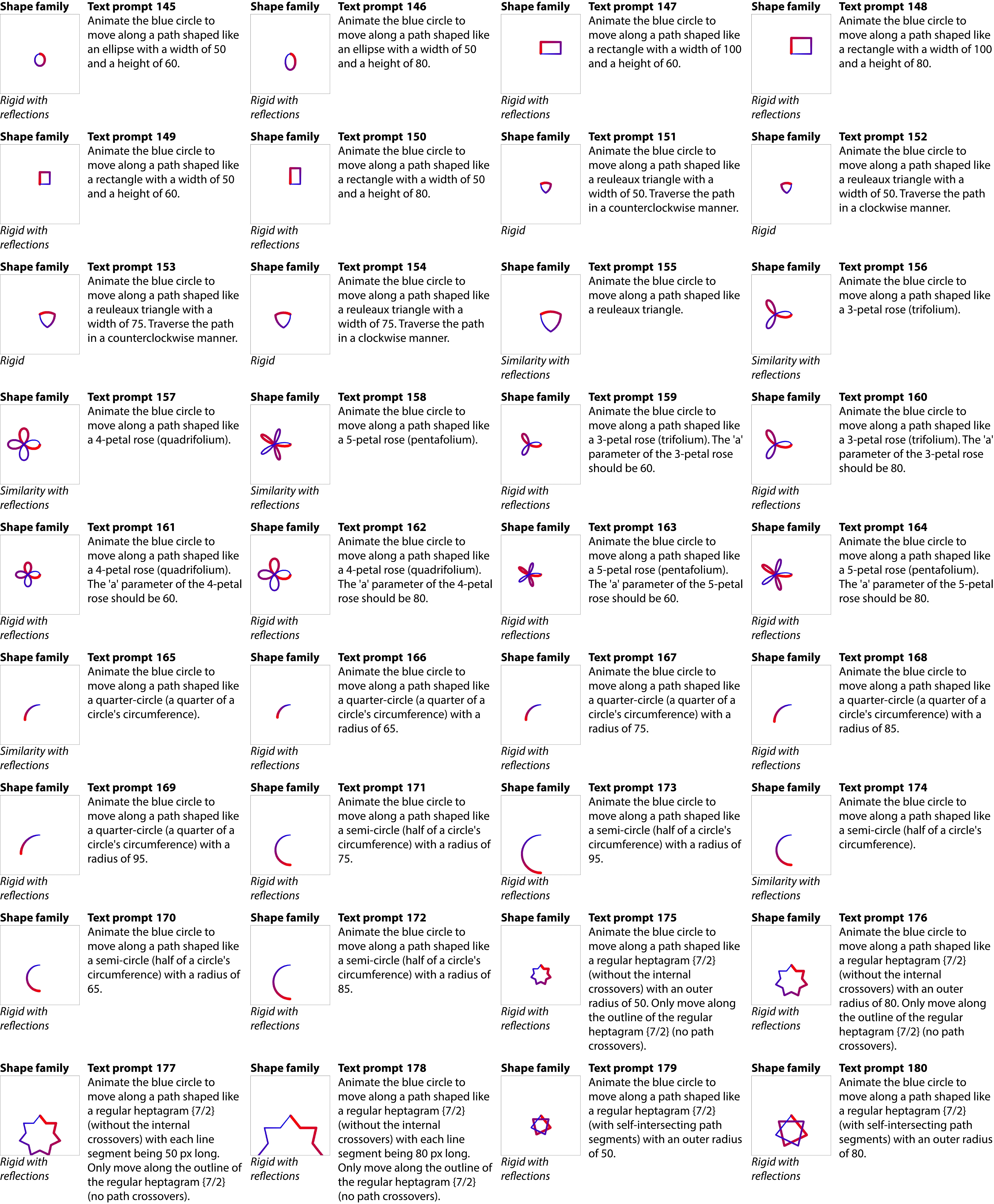}
    \caption{Motion trajectory benchmark: prompt 145--180 with ground truth shape families.}
    \label{fig:bench_mark_5}
\end{figure*}

\begin{figure*}[t]
    \centering
    \includegraphics[width=\textwidth]{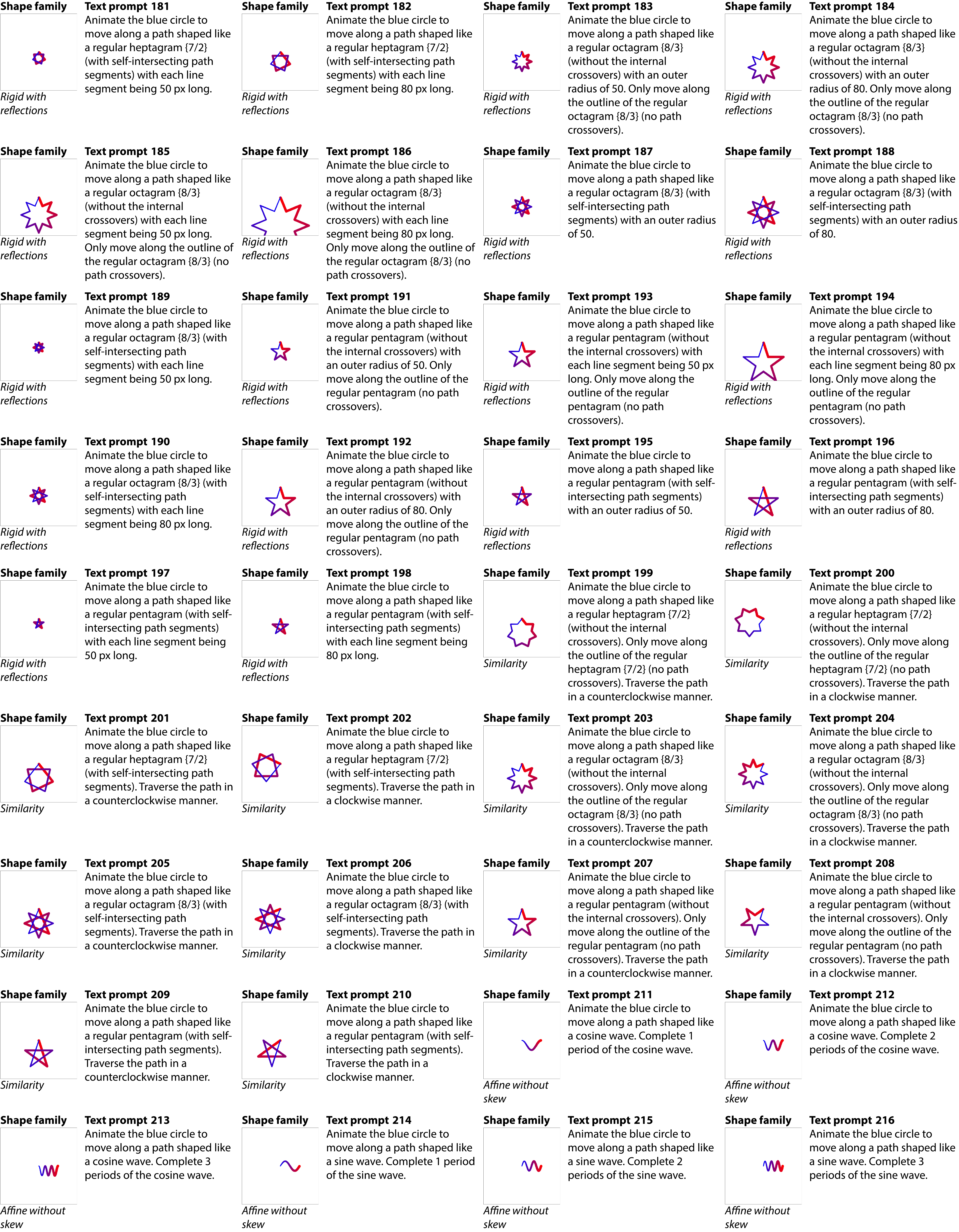}
    \caption{Motion trajectory benchmark: prompt 181--216 with ground truth shape families.}
    \label{fig:bench_mark_6}
\end{figure*}

\begin{figure*}[t!]
    \centering
    \includegraphics[width=\textwidth]{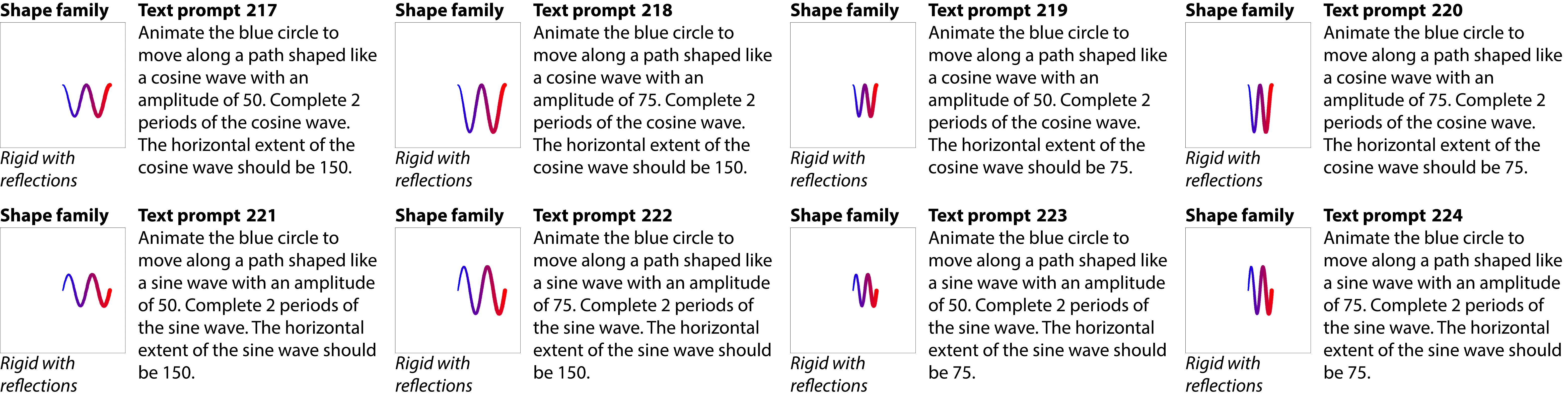}
    \caption{Motion trajectory benchmark: prompt 217--224 with ground truth shape families.}
    \label{fig:bench_mark_7}
\end{figure*}